\documentclass[APA,STIX2COL]{WileyNJD-v2}

\bibpunct{(}{)}{; }{}{}{}%
\usepackage{balance}
\usepackage{subfigure}
\usepackage{graphicx}
\usepackage{float}

\makeatletter
\def\NAT@aysep{,}
\makeatother

\articletype{Research Article}%

\begin{document}
\title{Sewer Image Super-Resolution with Depth Priors and Its Lightweight Network}

\author[1]{Gang Pan}

\author[1]{Chen Wang}

\author[1]{Zhijie Sui}

\author[2]{Shuai Guo}

\author[3]{Yaozhi Lv}

\author[4]{Honglie Li}

\author[5]{Di Sun*}

\author[6]{Zixia Xia*}

\authormark{Gang Pan \textsc{et al}}

\address[1]{\orgdiv{College of Intelligence and Computing}, \orgname{Tianjin University}, \orgaddress{\state{Tianjin}, \country{China}}}

\address[2]{\orgdiv{Department of Municipal Engineering}, \orgname{Hefei University of Technology}, \orgaddress{\state{Anhui}, \country{China}}}

\address[3]{\orgdiv{Key laboratory of Infrastructure Durability}, \orgname{Tianjin Municipal Engineering Design and Research Institute}, \orgaddress{\state{Tianjin}, \country{China}}}

\address[4]{\orgdiv{West-East Gas Pipeline Company}, \orgname{PipeChina}, \orgaddress{\state{Shanghai}, \country{China}}}

\address[5]{\orgdiv{College of Artificial Intelligence}, \orgname{Tianjin University of Science and Technology}, \orgaddress{\state{Tianjin}, \country{China}}}

\address[6]{\orgdiv{Donald Bren School of Information and Computer Sciences}, \orgname{University of California, Irvine}, \orgaddress{\state{California}, \country{USA}}}

\corres{Di Sun, College of Artificial Intelligence, Tianjin University of Science and Technology, No.1038 Dagu Nanlu, Tianjin 300222, China. \\
\email{dsun@tust.edu.cn} \\
\\
Zixia Xia, Donald Bren School of Information and Computer Sciences, University of California, Irvine, Donald Bren Hall, 6210, Irvine, CA 92697, USA. \\
\email{zixiax3@uci.edu}}

\abstract[Summary]{The Quick-view (QV) technique serves as a primary method for detecting defects within sewerage systems. However, the effectiveness of QV is impeded by the limited visual range of its hardware, resulting in suboptimal image quality for distant portions of the sewer network. Image super-resolution is an effective way to improve image quality and has been applied in a variety of scenes. However, research on super-resolution for sewer images remains considerably unexplored. 
In response, this study leverages the inherent depth relationships present within QV images and introduces a novel Depth-guided, Reference-based Super-Resolution framework denoted as DSRNet. It comprises two core components: a depth extraction module and a depth information matching module (DMM). DSRNet utilizes the adjacent frames of the low-resolution image as reference images and helps them recover texture information based on the correlation. By combining these modules, the integration of depth priors significantly enhances both visual quality and performance benchmarks.
Besides, in pursuit of computational efficiency and compactness, a super-resolution knowledge distillation model based on an attention mechanism is introduced. This mechanism facilitates the acquisition of feature similarity between a more complex teacher model and a streamlined student model, with the latter being a lightweight version of DSRNet.
Experimental results demonstrate that DSRNet significantly improves PSNR and SSIM compared with other methods. This study also conducts experiments on sewer defect semantic segmentation, object detection, and classification on the Pipe dataset and Sewer-ML dataset. Experiments show that the method can improve the performance of low-resolution sewer images in these tasks.} 


\maketitle

\section{Introduction}\label{sec1}

\begin{figure}[t]
    \centering
    \subfigure[ ]
    {\subfigure{\includegraphics[width=0.3\linewidth]{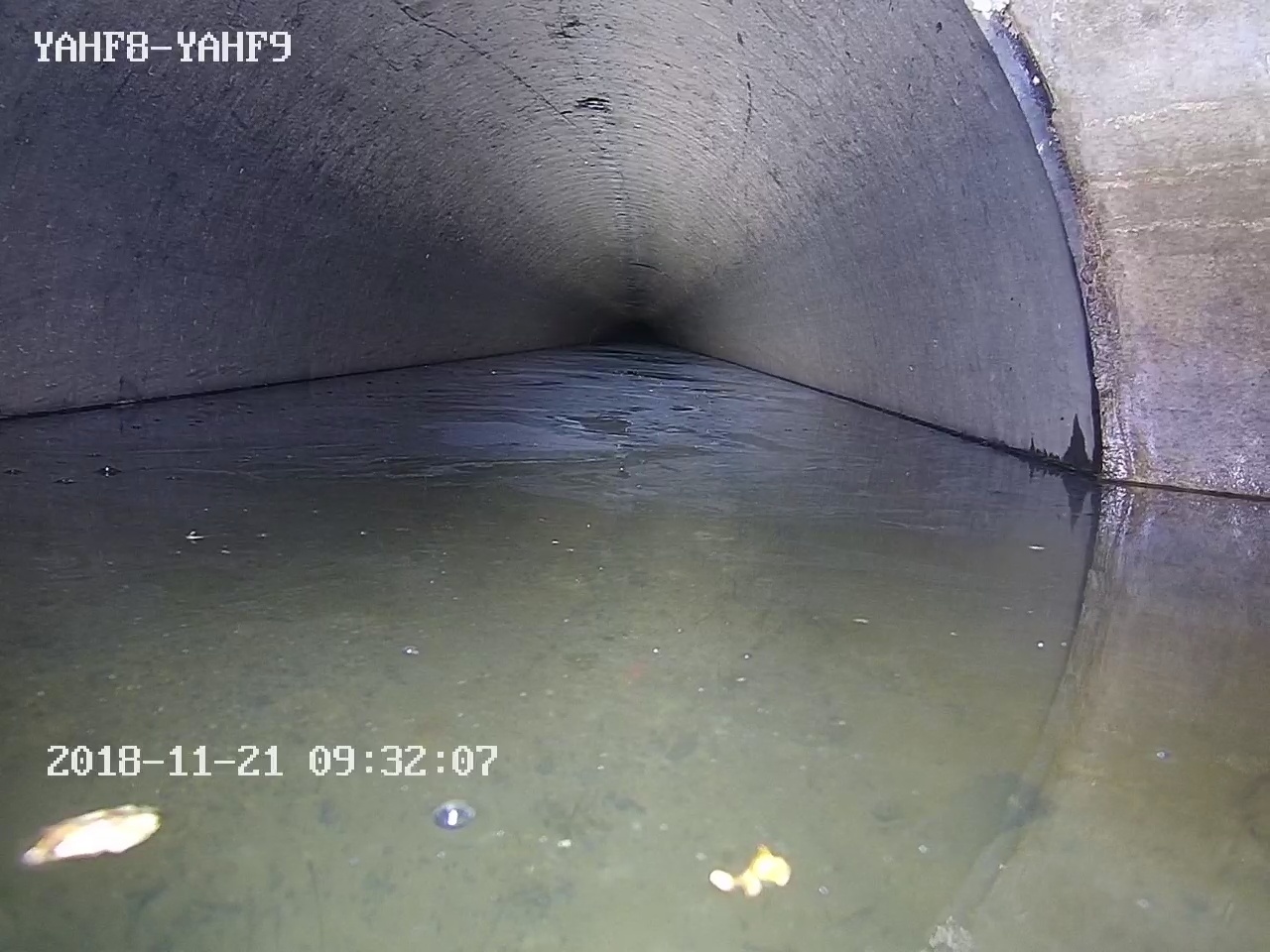}}
    \subfigure{\includegraphics[width=0.3\linewidth]{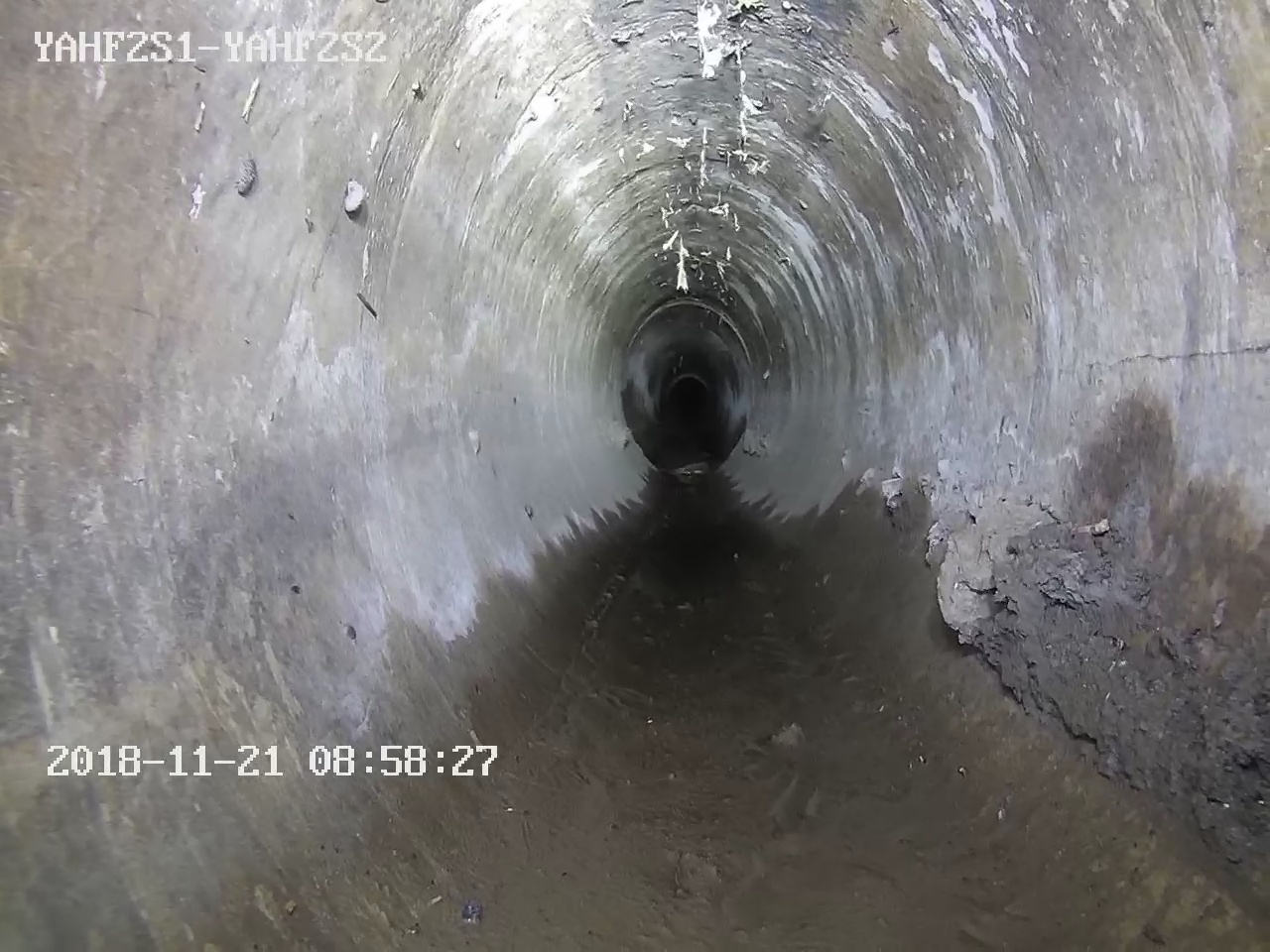}}
    \setcounter{subfigure}{0}
    \subfigure{\includegraphics[width=0.3\linewidth]{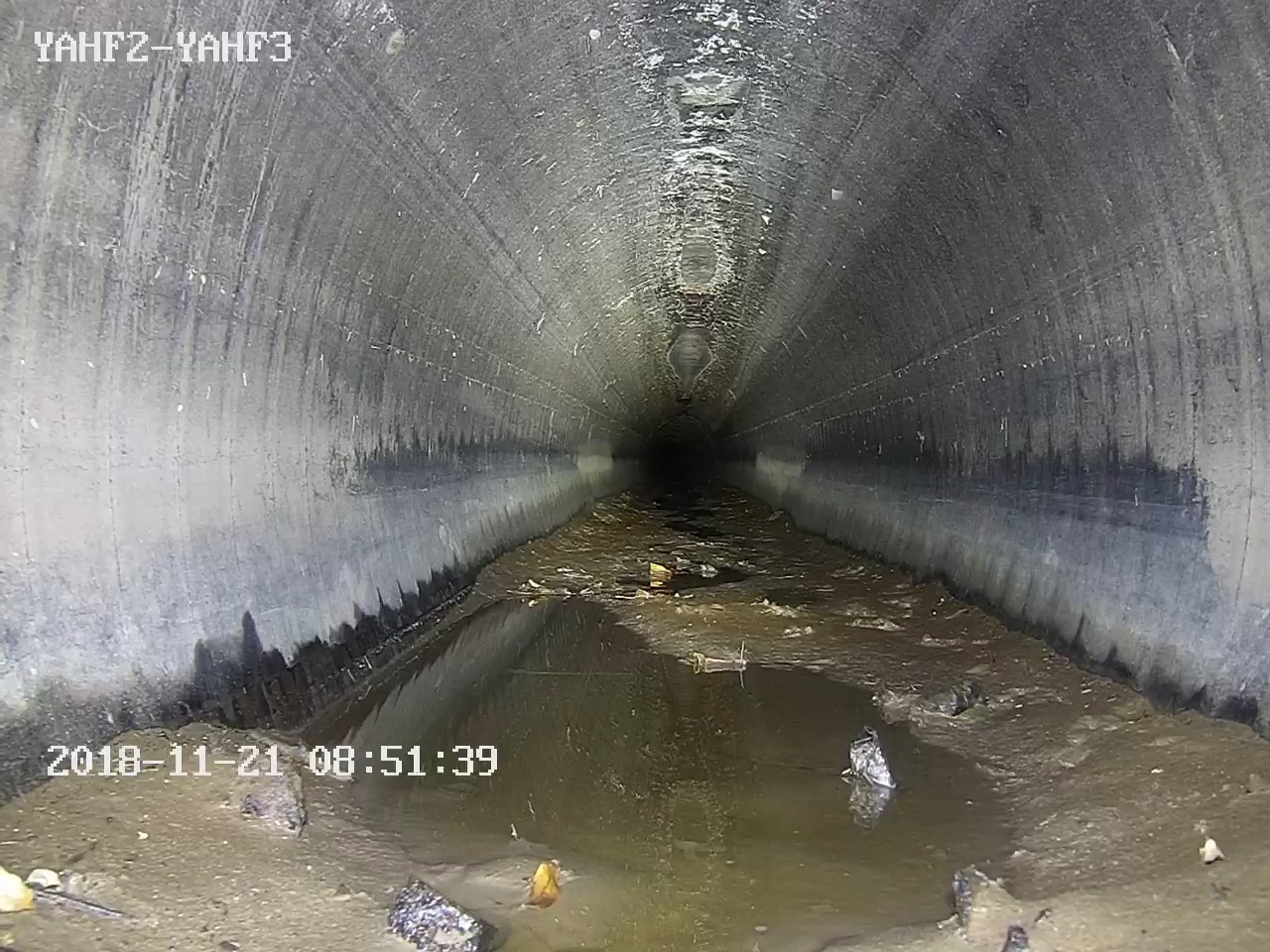}}}
    \subfigure[ ]
    {\subfigure{\includegraphics[width=0.3\linewidth]{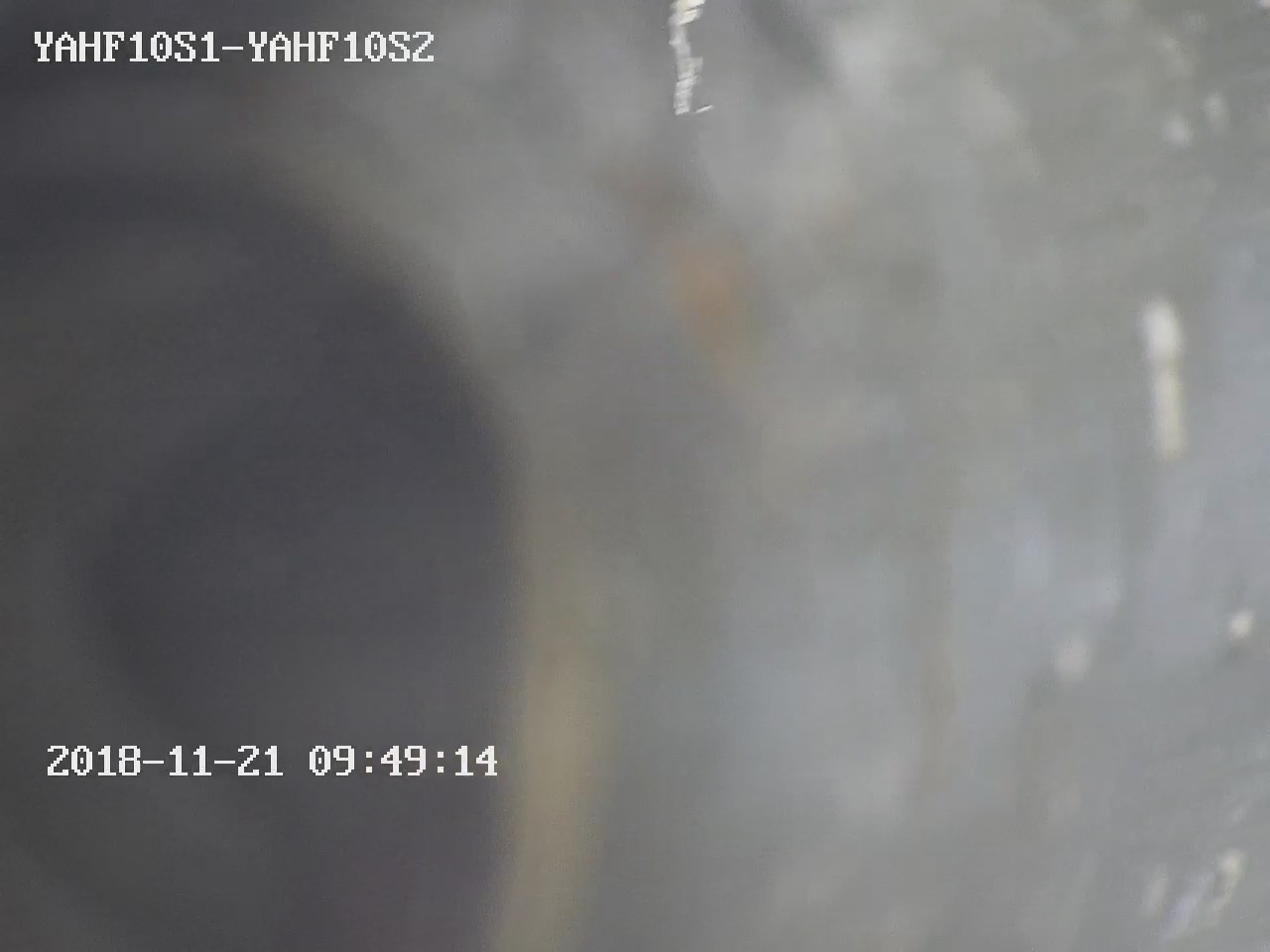}}
    \setcounter{subfigure}{0}
    \subfigure{\includegraphics[width=0.3\linewidth]{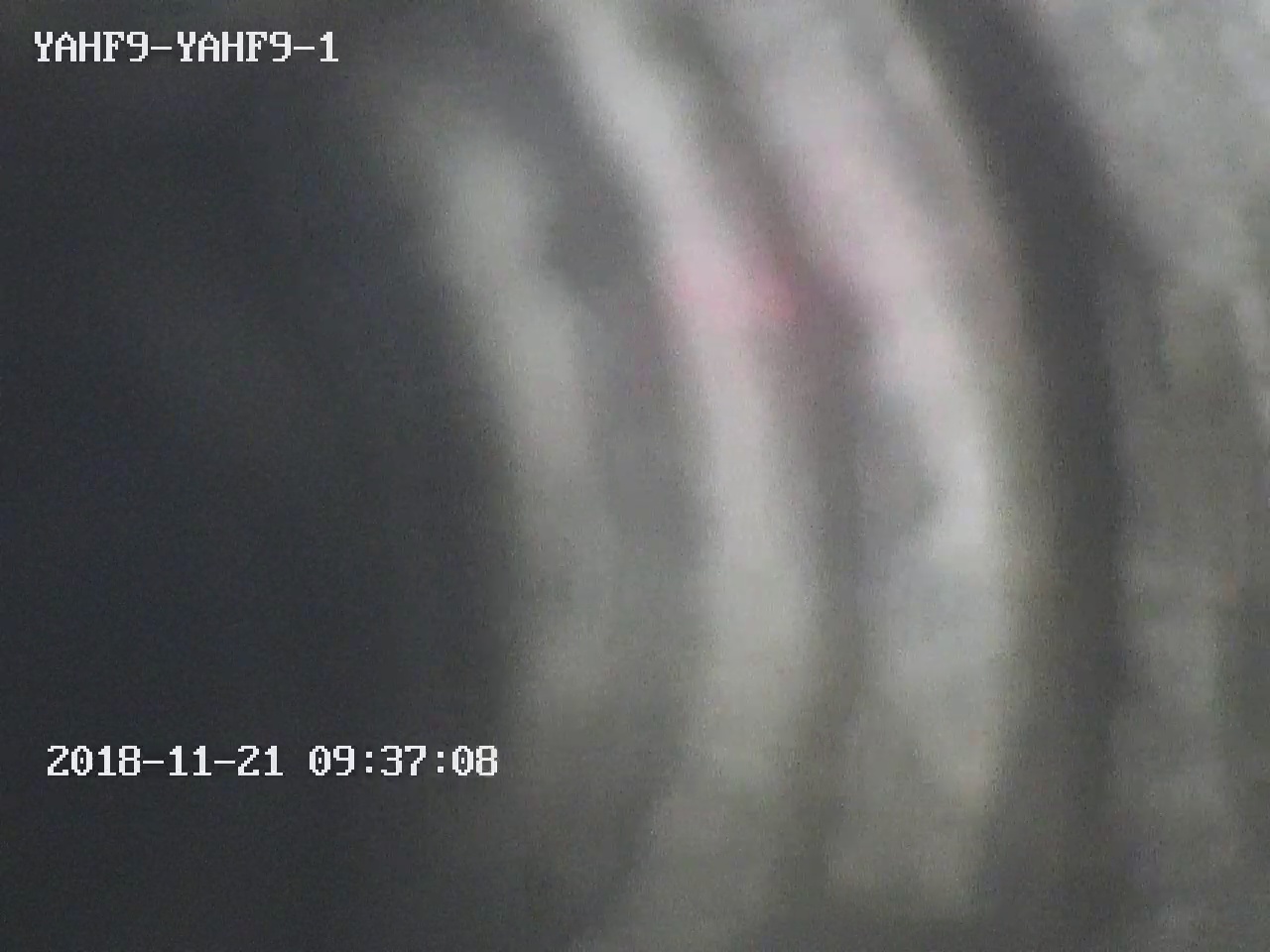}}
    \subfigure{\includegraphics[width=0.3\linewidth, height=56pt]{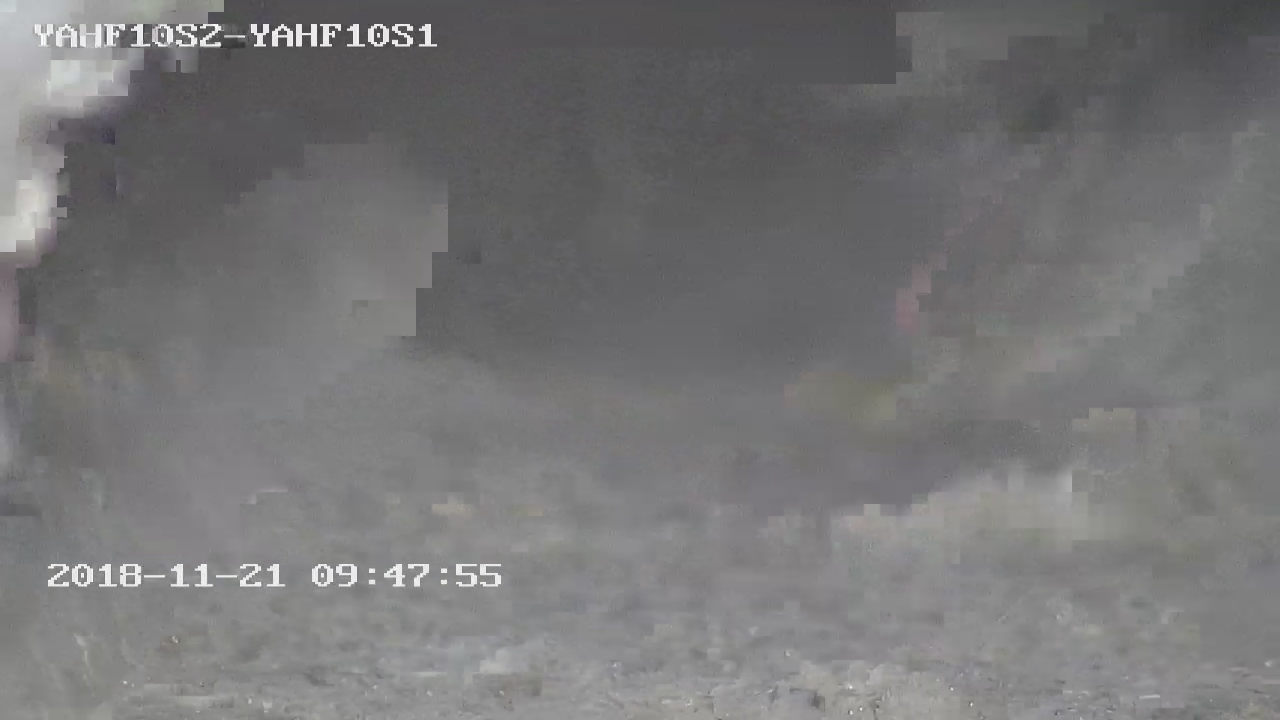}}}
    \caption{Images captured in sewer at different viewing distances. (a): Images at ideal viewing distances. (b): Images at far viewing distances. Image resolution is lower at far viewing distances }
    \label{mohu}
\end{figure}

The urban infrastructure relies extensively on the intricate web of sewer networks, signifying their indispensable role in sustaining the functionality of modern cities. Unfortunately, factors such as prolonged neglect and human interference can lead to the development of defects, including deformations, ruptures, and leaks. These challenges can severely disrupt the functionality of sewer system, posing risks to public safety. Therefore, regular defect detection of sewers has become a global consensus among municipal departments.
Currently, there are two primary methods for inspecting sewers: closed circuit television (CCTV) inspection and quick-view (QV) inspection. The latter, notable for its expediency and streamlined efficiency, has garnered preference. QV equipment captures real-time internal videos of sewers and swiftly records and saves images of defects. These images can undergo rigorous computer vision analyses, enabling accurate and reliable defect detection mechanisms. 
However, QV devices is inevitably constrained by both its visual coverage and the resolution capabilities of its imaging hardware. Consequently, the image resolution may decrease when the camera zooms in on remote sewer areas. Figure \ref{mohu} illustrates the potential appearance of internal sewer images at various viewing distances, highlighting that optimal line-of-sight areas yield high-resolution internal images of the sewer (as shown in Figure \ref{mohu}(a)). On the other hand, lines of sight that are excessively distant have lower resolutions (as shown in Figure \ref{mohu}(b)).
In real-world scenarios, subsequent computer vision tasks are typically carried out on captured QV images, including defect classification, target detection, semantic segmentation, etc. Nevertheless, low-resolution areas can hinder the seamless execution of these visual tasks, potentially causing accuracy to decrease. 
Super-resolution(SR) is a widely adopted technique utilized to enhance low-resolution images by restoring them to a higher resolution. This technique presents a promising avenue to handle the prevailing challenges outlined above. The QV device used for sewer inspection is an embedded mobile device capable of deploying the image super-resolution model. Large super-resolution models face challenges related to high deployment costs, slow detection speed, and insufficient flexibility on devices~\citep{kim2016accurate, lee2022local, kong2022reflash}.

As mentioned earlier, QV devices encounter the problem of localized low resolution at far viewing distances. Addressing this issue can significantly enhance the visibility within the sewer system, thereby optimizing the performance of subsequent computer vision tasks. For instance, enhancing the accuracy of sewer defect classification tasks can provide swift preliminary results for intelligent defect recognition software. Additionally, refining the precision of semantic segmentation tasks provides strong backing for computer-aided precise quantitative analysis conducted manually. To overcome the limited computational capabilities of QV devices, employing lightweight research models can effectively preserve device computing resources, improve operational efficiency, simultaneously maintain method accuracy.

At present, research on super-resolution in sewer inspection scenes remains insufficient. Most studies on image super-resolution focus on general-purpose image datasets~\citep{wang2020deep}, which lack experimentation on sewer image datasets. Additionally, there exists a geometric and textural correlation between sewer image sequences based on depth relationships. For super-resolution tasks, the depth information specific to sewer scenes can act as prior knowledge to constrain the search space for high-resolution image solutions. However, current general-purpose super-resolution algorithms only focus on reconstructing texture information~\citep{chen2023mffn} and do not take advantage of depth information.

To address these limitations, a method for sewer image super-resolution based on depth prior and reference images is proposed. A depth extraction network is incorporated into the original super-resolution network and trained together with it. This allows the integration of depth information from low-resolution images and reference images, enhancing the performance of the image super-resolution network by generating a depth map. 

A depth information matching module is designed to further enhance the accuracy of the super-resolution algorithm in restoring textures. After extracting features from the depth map, it is then combined with the feature maps of the low-resolution image and reference image. This restricts the solution space during the block matching process between low-resolution and reference images, emphasizing high resolution.

Furthermore, a super-resolution knowledge distillation model based on attention feature matching is also designed. The model applies knowledge distillation to the super-resolution model, introducing an attention mechanism into the feature selection process of both the teacher model and the student model. A new loss function is proposed specifically for image super-resolution tasks.  By employing knowledge distillation, the size of the model is reduced while preserving the effectiveness of the algorithm. This approach achieves model compression and acceleration, which is beneficial for deploying the model.

\section{Related Work}\label{sec2}

\subsection{Super-resolution}

Currently, there are some neural networks designed for defect detection in sewer pipelines such as the work by \cite{li2023attention} and \cite{ma2023transformer}. However, most of these networks are tailored for CCTV inspections and lack specific methods for handling low-resolution scenarios. Based on the required types of training data, image super-resolution methods can be broadly categorized into single image super-resolution (SISR) and reference-based super-resolution (RefSR). SISR methods~\citep{wang2024sinsr, garber2024image} are using transformer-based models or diffusion-based models. Recently, \cite{ray2024cfat, zhang2024transcending} put more focus on shifted window technique, aiming to find the more general approach for restroing images. Compared to single image super-resolution methods, reference-based super-resolution methods utilize additional reference images to perform image upsampling operations on low-resolution input images. Reference images frequently share similar content with low-resolution images and can provide high-frequency details, significantly enhancing the super-resolution recovery process. In recent years, most super-resolution methods based on reference images have been constructed using convolutional neural networks. One of the prevailing approaches in this domain involves conducting spatial alignment operations between reference images and low-resolution images~\citep{zheng2018crossnet, shim2020robust, zhu2019deformable}. This method primarily utilizes optical flow to guide the matching process between reference images and low-resolution images. Another common approach is to utilize image block matching~\citep{zhang2019image, yang2020learning, xia2022coarse}. By dividing the image into different image blocks, the similarity between the blocks to match the low-resolution image with the reference image can be utilized. \cite{chu2023implicit} also attempted to apply super-resolution technology to crack detection. However, their network only incorporates two key enhancements of super-resolution and still relies on high-definition camera inputs, making it unsuitable for QV devices. 

\subsection{Knowledge distillation}

\cite{gou2021knowledge} highlighted that deep learning has achieved remarkable success in both academia and industry because of its scalability and ability to encode large-scale data. However, deploying deep neural networks on embedded mobile devices is frequently constrained by hardware limitations. This has spurred the development of numerous model compression and acceleration techniques, with knowledge distillation being particularly notable.

\cite{hinton2015distilling} initially proposed this architecture and labeled the technique of exploiting it as "knowledge distillation." This approach first utilizes the teacher network to compute the "soft target" and then incorporates both the soft label and the actual label as the ground truth during the training of the student network. The outcomes demonstrate that using this training strategy can significantly improve the image classification performance in the MNIST dataset~\citep{lecun1998gradient}. Subsequently, scholars have proposed various methods of distillation, including mutual distillation~\citep{zhang2018deep} and self-distillation~\citep{zhang2019your}, which have broadened the research scope and application scenarios for knowledge distillation.

\subsection{Depth estimation}

Depth estimation is a fundamental problem in computer vision, with wide range applications in fields such as 3D reconstruction, virtual reality, and autonomous driving. Monocular depth estimation refers to utilizing only one color image to calculate the distance between the camera and each pixel. Scholars have undertaken extensive research into depth estimation methods utilizing convolutional neural networks. CLIFFNet~\citep{wang2020cliffnet} employs a multi-scale fusion convolutional framework to improve the quality of depth estimation prediction results. SANL-Net~\citep{xia2023structure} adapts monocular depth cues for pipe defects detection with CCTV images, innovatively uses water-pipewall borderlines to guide the dehazing dataset production. This previous work has inspired us to apply depth information to super-resulotion tasks for QV detection. The monodepth2~\citep{godard2019digging} is an excellent method for depth estimation. The innovative points of this method are as follows: (1) Proposing an automatic masking method that can effectively ignore unobservable pixels during monocular training. (2) Introducing an appearance-matching loss to address the issue of occluding pixels during monocular supervision. (3) Proposing a multi-scale appearance matching loss by rescaling all images to the resolution of the input image for calculation. This approach helps to reduce artifacts in the generated outcomes. In recent years, an increasing number of researchers have embarked on studying depth estimation based on transformer. Building on the triumph of employing transformer in computer vision tasks, TransDepth~\citep{zhao2021transformer} and DPT~\citep{ranftl2021vision} have enhanced the performance of the model by replacing convolution operations with transformer network layers. 

\begin{figure*}[t]
    \centering
    \includegraphics[width=0.9\textwidth]{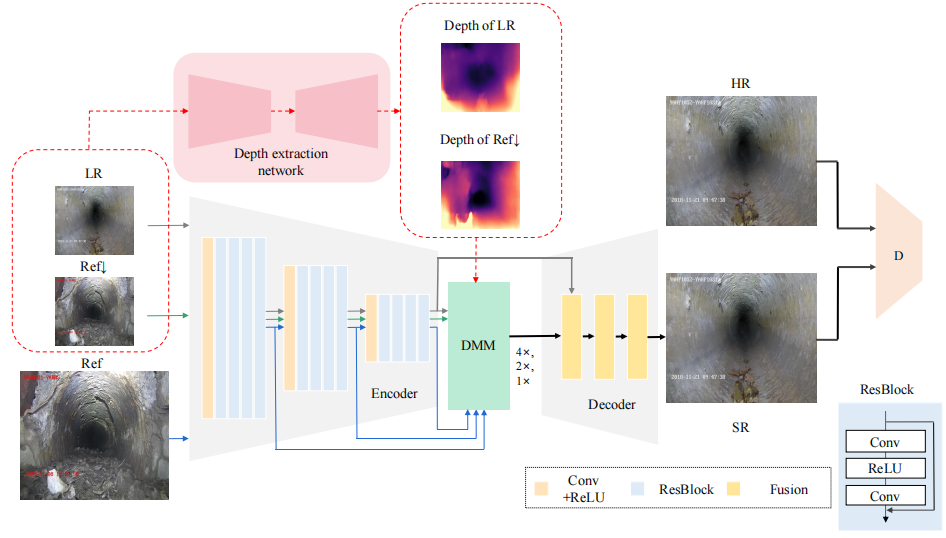}
    \caption{The overall architecture of proposed DSRNet contains five module: depth extraction network, encoder, depth matching module (DMM), decoder, and discriminator. LR, Ref$\downarrow$, Ref, HR, and SR represent low-resolution images, downsampled reference images, reference images, high-resolution images and super-resolution images}
    \label{DSRNet}
\end{figure*}

\section{Depth and reference based Super-Resolution Network}\label{sec3}

This paper proposes a novel Depth and Reference Based Super-Resolution Network (DSRNet) that uses depth priors and reference images. DSRNet comprises the following modules: a depth extraction network, an encoder, a depth matching module (DMM), a decoder, and a discriminator. Let $I_{LR}$ denote the input low-resolution image, $I_{Ref}$ denote the input reference image, and $I_{Ref\downarrow}$ denote the downsampled reference images, where the downsample scale $S$ is 4. The function of the depth extraction network is to generate the depth map of $I_{LR}$, with $I_{Ref\downarrow}$ serving as an auxiliary input for subsequent modules to use. These two depth maps are represented respectively by $D_{LR}$ and $D_{Ref\downarrow}$. The encoder's task is to extract features from $I_{LR}$, $I_{Ref}$, and $I_{Ref\downarrow}$. The encoder consists of three feature extraction modules. After feeding the images into the encoder, three different scale features are obtained, denoted as $F^s_{LR}$, $F^s_{Ref\downarrow}$, and $F^s_{Ref}$, where $s$ takes values of 1, 2, or 4, representing the scaling scale. Subsequently, $F^s_{Ref}$, $F^s_{LR}$, $F^s_{Ref\downarrow}$, $D_{LR}$, and $D_{Ref\downarrow}$ are collectively inputted into the DMM. The DMM extracts depth map features and integrates them with the features extracted by the encoder, enabling feature matching between low-resolution and reference images. The decoder converts the output processed by DMM $F^s_{Ref}$ and $F^s_{LR}$ into a high-resolution image. Finally, the discriminator determines the authenticity of the network's output results and improves the quality of the image super-resolution method as much as possible. The overall network model structure is shown in Figure \ref{DSRNet}. Pairwise LR-HR images are required for network training. In the case of lack of pairwise LR-HR images, the idea of self-supervised learning(SSL) can be adopted to directly downsample HR as LR, and then try to restore LR to HR~\citep{rafiei2022self}.

\subsection{Depth extraction network}

The depth extraction network plays a pivotal role in generating depth maps of low-resolution and downsampled reference images, thereby facilitating the utilization of depth information in sewer scenes. The U-Net structure, which is an encoder-decoder network with skip connections, is adopted for this purpose. The entire network process is denoted by $U$. This structure not only extracts depth features but also effectively preserves the original information. The input of the depth extraction network includes a low-resolution image and a downsampled reference image, represented by $I_{LR}$ and $I_{Ref\downarrow}$, respectively. The output of the depth extraction network is represented by $D_{LR}$ and $D_{Ref\downarrow}$.

\subsection{DMM}

\begin{figure*}[t]
    \centering
    \includegraphics[width=0.7\textwidth]{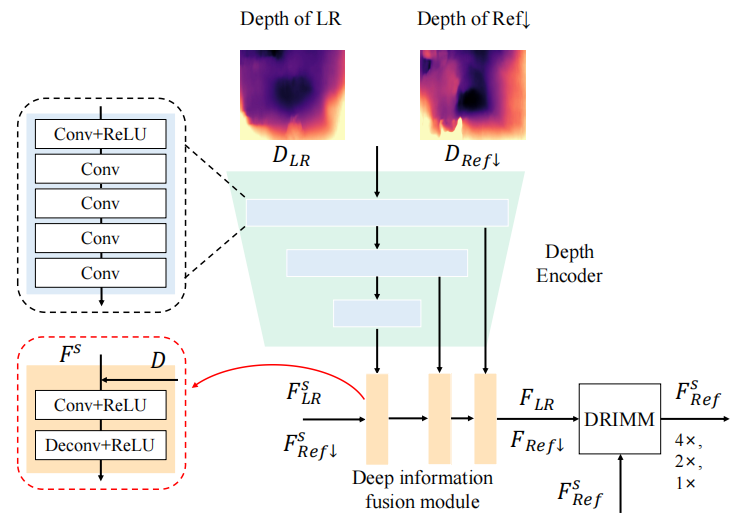}
    \caption{The architecture of DMM which contains three main modules: depth encoder, depth information fusion module and depth-based reference image matching module}
    \label{dmm}
\end{figure*}

The DMM plays a critical role in extracting depth map features, merging them with the features extracted by the encoder, and conducting feature matching operations. The structure of DMM is depicted in Figure \ref{dmm} and primarily includes a depth encoder, a depth information integration module, and a Depth-based Reference Image Matching Module (DRIMM).

The specific process of DMM involves firstly inputting $D_{LR}$ and $D_{Ref\downarrow}$ into the depth encoder to obtain depth map features of different scales, represented by $D^s_{LR}$ and $D^s_{Ref\downarrow}$. The output of the encoder is then adjusted to $F^s_{LR}$ and $F^s_{Ref\downarrow}$, and the output of the depth encoder $D^s_{LR}$ and $D^s_{Ref\downarrow}$ is transmitted to the depth information integration module through connection operations. The output of the depth information integration module is $F_{LR}$ and $F_{Ref\downarrow}$, and these two features are combined with $F^s_{Ref}$ in DRIMM.

\paragraph{Depth Encoder}
The depth encoder is responsible for extracting depth map information from the inputs $D_{LR}$ and $D_{Ref\downarrow}$. The outputs are represented by depth map features $D^s_{LR}$ and $D^s_{Ref\downarrow}$. The structure of the depth encoder is shown in Table \ref{ste}, which consists of three feature extraction modules, with each including a convolution layer and four residual blocks. However, unlike RGB images, depth images have a single-channel input. Hence, the first convolution layer in this section has 1 input channel instead of 3. The depth encoder is denoted as $E_D$, and the process of the depth encoder network can be formalized as Equation (\ref{D1}) and Equation (\ref{D2}).
\begin{equation}
    D^s_{LR} = E_d(D_{LR})
    \label{D1}
\end{equation}
\begin{equation}
    D^s_{Ref\downarrow} = E_d(D_{Ref})
    \label{D2}
\end{equation}

\paragraph{Depth Information Fusion Module}
The depth information fusion module comprises three sets of convolution and deconvolution layers. Each set consists of a $3\times3$ convolution layer with step size 1 and a $3\times3$ deconvolution layer with step size 2. After each convolution and deconvolution operation, the ReLU activation function is applied for processing. The purpose of this module is to fuse the features of $D^s$ and $F^s$. It takes inputs of $D^s$ and $F^s$, and produces outputs of fused features of low-resolution and downsampled reference images, represented as $F_{LR}$ and $F_{Ref\downarrow}$, respectively. Define $\oplus$ as the connection operation, the process of the depth information fusion module can be formalized as Equation (\ref{M1}) and Equation (\ref{M2}).
\begin{equation}
    F_{LR} = Deconv(Conv(D^s_{LR}{\oplus}F^s_{LR}))
    \label{M1}
\end{equation}
\begin{equation}
    F_{Ref\downarrow} = Deconv(Conv(D^s_{Ref\downarrow}{\oplus}F^s_{Ref\downarrow}))
    \label{M2}
\end{equation}

\paragraph{Depth-based Reference Image Matching Module}
 
\begin{equation}
r^k_{c,j} = <\dfrac{p^k_c}{||p^k_c||},\dfrac{q_j}{||q_j||}>
\label{rcos}
\end{equation}

where $p^k_c$ denotes the central image area of $B^k_{LR}$, $q_j$ represents the $j$-th image area of $F_{Ref\downarrow}$, and $r^k_{c,j}$ is the similarity score between $p^k_c$ and $q_j$. Based on this, this work crop an image block of size $d_x{\times}d_y$ in $F_{Ref\downarrow}$ centered around the image area with the highest similarity to $p^k_c$, denoted by $B^k_{Ref\downarrow}$, where $d_x$ and $d_y$ represent the length and width of the image block. Simultaneously, an image block of size $sd_x{\times}sd_y$ from $F^s_{Ref}$, denoted by $B^{s,k}_{Ref}$ is cropped. Afterward, a triplet $(B^k_{LR}, B^k_{Ref\downarrow}, B^{s,k}_{Ref})$ is obtained.

The next step is matching image area for each $B^k_{LR}$ and $B^k_{Ref\downarrow}$ to obtain a set of index maps $\{D^0, ..., D^{K-1}\}$ and similarity score maps $\{R^0, ..., R^{K-1}\}$. For the $k$-th binary $(B^k_{LR}, B^k_{Ref\downarrow})$, let $p^k_i$ denote the $i$-th image area of $B^k_{LR}$, $q^k_j$ be the $j$-th image area of $B^k_{Ref\downarrow}$, and $r^k_{i,j}$ denote the similarity score between $p^k_i$ and $q^k_j$. The calculation of the $i$-th element of index map $N^k$ is shown in Equation (\ref{Nki}).
\begin{equation}
N^k_i = \underset{j}{\arg \max} r^k_{i,j}
\label{Nki}
\end{equation}
The $i$-th element of the similarity score map $R^k_i$ represents the highest similarity score between the $i$-th image area associated with $B^k_{LR}$ and all image areas in $B^k_{Ref\downarrow}$. This value is computed using Equation (\ref{Rki}).
\begin{equation}
R^k_i = \underset{j}{\max} r^k_{i,j}
\label{Rki}
\end{equation}

Based on the index map $N^k$, image areas are extracted from the image block $B^{s,k}_{Ref}$ and form a new feature map denoted by $B^{s,k}_M$. Specifically, set the $N^k_i$-th image area of $B^{s,k}_{Ref}$ as the $i$-th image area of $B^{s,k}_M$, and then multiply $B^{s,k}_M$ with the corresponding similarity score mapping $R^k$ to obtain a weighted feature image block. The calculation formula for the weighted feature image block $W^{s,k}_M$ is given by Equation (\ref{BM}).
\begin{equation}
W^{s,k}_M = B^{s,k}_M {\odot} (R^k){\uparrow}
\label{BM}
\end{equation}
where $(){\uparrow}$ represents bicubic interpolation, and $\odot$ denotes element-wise multiplication. The final output of DRIMM is the folded $\{W^{s,0}_M, ..., W^{s,K-1}_M\}$, which is represented as $F^s_{Ref}$. For each image block in $F_{LR}$, the image block in the corresponding position in the matched $F^s_{Ref}$ is the most similar one in the original $F^s_{Ref}$.

\subsection{Loss functions}

\textbf{Depth Reconstruction Loss.} For the depth extraction network, this paper employs the $L_1$ loss function to calculate the difference between corresponding pixels in the two generated depth maps and the two depth ground truths, and subsequently sums them. The overall depth reconstruction loss of the entire depth extraction network is computed as shown in Equation (\ref{ldep}).
\begin{equation}
    L_{dep} = ||\tilde{D}_{Ref\downarrow}-D_{Ref\downarrow}||_1 + ||\tilde{D}_{LR}-D_{LR}||_1
    \label{ldep}
\end{equation}
where $D_{LR}$ and $D_{Ref\downarrow}$ denote the low-resolution depth map output and downsampled reference depth map output of the depth extraction network, respectively. Meanwhile, $\tilde{D}_{LR}$ and $\tilde{D}_{Ref\downarrow}$ represent the depth ground truth for the low-resolution image and downsampled reference image, respectively.

\textbf{Reconstruction Loss.} The reconstruction loss function aims to minimize the pixel-wise distance between the network output and the ground truth, thereby reducing their differences. In this paper, an $L_1$ loss function is adopted as the reconstruction loss function, which is expressed as shown in Equation (\ref{lrec}).
\begin{equation}
    L_{rec}=||I_{HR}-I_{SR}||_1
    \label{lrec}
\end{equation}
where $I_{HR}$ and $I_{SR}$ denote the ground truth and the network output results, respectively.

\textbf{Perception Loss.} Relying solely on pixel-wise loss is inadequate in capturing the visual quality of an image as perceived by the human eye, especially concerning the accuracy of detail recovery. Extracting features using VGG and other networks can be helpful in restoring such details. Perception loss involves computing the differences between the network output and ground truth after feature extraction using the VGG network. The formula for the perception loss function is provided in Equation (\ref{lper}).
\begin{equation}
    L_{per}=||\phi_{i}(I_{HR})-\phi_{i}(I_{SR})||_2
    \label{lper}
\end{equation}
where $\phi_{i}$ represents VGG19.

\textbf{Adversarial Loss.} As described earlier, the discriminator introduced in this paper employs an adversarial loss function $L_{adv}$ to distinguish between authentic and generated images, which is advantageous in improving the visual quality and naturalness of generated images. Specifically, the network utilizes the loss function of relative discriminators (Relative GANs). If $D$ denotes the discriminator, then the computation of $L_{adv}$ is given by Equation (\ref{lg}), Equation (\ref{ld}), and Equation (\ref{ladv}).
\begin{equation}
    L_{G}=-\mathbb{E}_{I_{HR}}[log(1-D(I_{HR},I_{SR}))]-\mathbb{E}_{I_{SR}}[log(D(I_{SR},I_{HR}))]
    \label{lg}
\end{equation}
\begin{equation}
    L_{D}=-\mathbb{E}_{I_{HR}}[log(D(I_{HR},I_{SR}))]-\mathbb{E}_{I_{SR}}[log(1-D(I_{SR},I_{HR}))]
    \label{ld}
\end{equation}
\begin{equation}
    L_{adv}=\lambda_{G}L_{G}+\lambda_{D}L_{D}
    \label{ladv}
\end{equation}
where $\lambda_{G}$ and $\lambda_{D}$ represent the weight parameters for the generator loss and discriminator loss, respectively.

\begin{figure*}[t]
    \centering
    \includegraphics[width=0.9\textwidth]{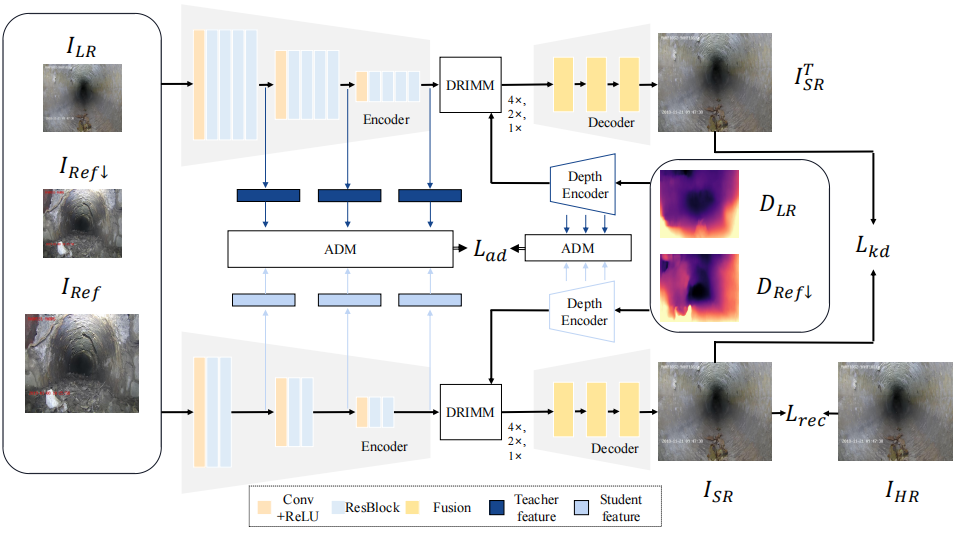}
    \caption{The overall architecture of knowledge distillation based on attention feature matching. The model uses DSRNet called DSRNet-T as the teacher model, and a lightweight DSRNet called DSRNet-S as student model. The Attention-based Distillation Module (ADM) can effectively selects, matches, and distills the essential features from the teacher model and the student model}
    \label{disnet}
\end{figure*}

\textbf{Overall Loss.} As different loss functions serve different purposes, they need to be assigned appropriate weights to balance their contributions. Specifically, $\lambda_{dep}$, $\lambda_{rec}$, $\lambda_{per}$, and $\lambda_{adv}$ denote weight parameters used to balance the proportions of the depth reconstruction loss, reconstruction loss, perception loss, and adversarial loss, respectively. The complete loss function of the network is expressed as shown in Equation (\ref{l1}).
\begin{equation}
    L=\lambda_{dep}L_{dep}+\lambda_{rec}L_{rec}+\lambda_{per}L_{per}+\lambda_{adv}L_{adv}
    \label{l1}
\end{equation}

\section{Knowledge distillation based on attention feature matching}\label{sec4}

This paper presents an innovative approach to super-resolution knowledge distillation, which utilizes attention feature matching to compress the network model, as showed in Figure \ref{disnet}. The proposed method uses DSRNet as the teacher model, denoted as DSRNet-T, and a lightweight version of DSRNet, called DSRNet-S as the student model. Table \ref{ste} depicts the architecture of both models. To facilitate knowledge transfer between the teacher and student models, this work designs an Attention-based Distillation Module (ADM), which effectively selects, matches, and distills the essential features from the teacher model and the student model. In the following sections, the teacher model, student model, and feature distillation module of our proposed network will be introduced.

\begin{table}[!h]
    \centering
    \caption{Encoder and depth encoder structure of DSRNet-T and DSRNet-S}
    \begin{tabular}{cc|cc}
    \toprule[1.5pt]
        \multicolumn{2}{c|}{DSRNet-T} & \multicolumn{2}{c}{DSRNet-S} \\
        \midrule[0.5pt]
        Sturct & Number & Struct & Number \\ 
        \midrule[0.5pt]
        Conv & 1 & Conv & 1 \\
        ResBlock & 4 & ResBlock & 2 \\
        Conv & 1 & Conv & 1 \\
        ResBlock & 4 & ResBlock & 2 \\
        Conv & 1 & Conv & 1 \\
        ResBlock & 4 & ResBlock & 2 \\
        \bottomrule[1.5pt]
    \end{tabular}
    \label{ste}
\end{table}

\begin{figure*}[!h]
    \centering
    \includegraphics[width=0.9\textwidth]{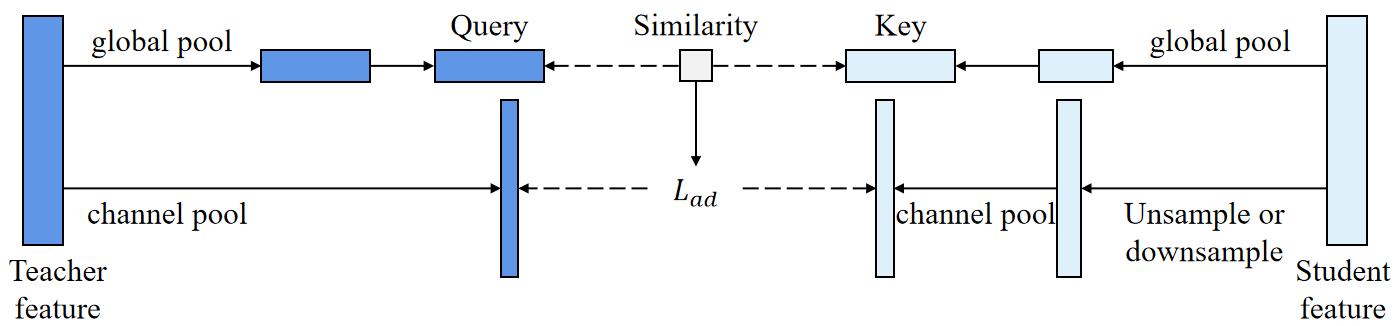}
    \caption{Attention-based distillation module(ADM)}
    \label{ad}
\end{figure*}

\subsection{Attention-based distillation module}

In Attention-based distillation module, the teacher model serves as the source of knowledge. The student model is designed to reach the performance of the teacher model as closely as possible under limited computational resources. The attention mechanism calculates the similarity between teacher and student features using query and key vectors, which will guide the learning process. Figure \ref{ad} illustrates the structure and principle of our proposed Attention-based Distillation Module (ADM). Specifically, take individual teacher features and student features as examples. Firstly, the features are processed through global average pooling and channel pooling operations, respectively. Next, the globally pooled teacher features serve as query vectors (Query, Q), while the globally pooled student features are utilized as key vectors (Key, K). The similarity between teacher and student features is then calculated using these two vectors, and an $L2$ loss function is further applied to the pooled channel teacher and student features. Finally, the similarity is used as a weighted value to multiply the loss, resulting in the attention distillation loss for a single teacher and a single student feature. Our feature distillation module aims to minimize the sum of attention distillation losses between all teacher features and all student feature binaries, promoting effective knowledge transfer from the teacher the to student features.

Once the input passes through the encoder, a set of teacher middle layer features $f^T$ and a set of student middle layer features $f^S$ are obtained, where $f^T = \{ f^T_1, f^T_2, ..., f^T_M\}$ and $f^S = \{ f^S_1, f^S_2, ..., f^S_N\}$. Here, $M$ and $N$ denote the number of teacher and student features, respectively. If $H$, $W$, and $c$ represent the height, width, and number of channels of the feature map, then the size of each feature map can be expressed as $H {\times}W{\times}c$. The feature distillation module takes $f^T$ and $f^S$ as inputs, with the goal of aligning these two sets of features and calculating the similarity of all teacher-student feature pairs to transfer the knowledge of teacher features to the corresponding student features.

Let $f^T_M$ and $f^S_N$ denote the $m$-th and $n$-th features of $f^T$ and $f^S$, respectively. The attention mechanism by \cite{xu2015show} is used to compute their similarity. The first step is to generate the query vector $\textbf{q}_{m}$ from $f^T_m$ and key vector $\textbf{k}_{n}$ from $f^S_n$. The calculation process of $\textbf{q}_{m}$ and $\textbf{k}_{n}$ is presented in Equation (\ref{qm}) and Equation (\ref{kn}).
\begin{equation}
    \textbf{q}_m = g_Q(W^Q_m\cdot{\phi}^{HW}(f^T_m))
    \label{qm}
\end{equation}
\begin{equation}
    \textbf{k}_n = g_K(W^K_n\cdot{\phi}^{HW}(f^S_n))
    \label{kn}
\end{equation}
where ${\phi}^{HW}(\cdot)$ denotes the global average pooling operation, and $g_Q$ and $g_K$ are the activation functions of the query vector and key vector, respectively. The matrices $W^Q_m\in\mathbb{R}^{c{\times}c^T_m}$ and $W^K_n\in\mathbb{R}^{c{\times}c^S_n}$ represent the linear transformation matrices for the $m$-th query vector and the $n$-th key vector, respectively. Since the features of $f^T_m$ and $f^S_n$ come from different parts of the network and convey different information, the convolution layers are employed to implement the linear transformation matrix for each feature.

Once the query and key vectors are computed, the softmax activation function is used to calculate the attention value ${\alpha}_{m,n}$ between the teacher feature and student feature. This value provides guidance for the feature selection process as a measure of the similarity between the teacher and student features. The calculation formula for ${\alpha}_{m,n}$ is presented in Equation (\ref{alphamn}).
\begin{equation}
    {\alpha}_{m,n} = \text{softmax}((\textbf{q}^{\top}_{m}W^{Q-K}_{n}\textbf{k}_{n}+(\textbf{p}^T_m)^{\top}\textbf{p}^S_n)/\sqrt{c})
    \label{alphamn}
\end{equation}
where $W^{Q-K}_{n}\in\mathbb{R}^{c{\times}c}$ denotes the weight conversion matrix used to transform the dimensions of vectors for calculation. In our work, $W^{Q-K}_{n}$ represents a convolution layer. $\textbf{p}^T_m\in\mathbb{R}^{c}$ and $\textbf{p}^S_n\in\mathbb{R}^{c}$ represent the position embeddings, which add positional information to features. The attention value ${\alpha}_{m,n}$ measures the similarity between the $m$-th teacher feature and the $n$-th student feature. Therefore, the attention value between the $m$-th teacher feature and all student features can be expressed as ${\alpha}_m = \{{\alpha}_{m,1}, {\alpha}_{m,2}, ..., {\alpha}_{m,N}\}$. By utilizing ${\alpha}_m$, the student model can selectively learn the teacher feature $f^T_m$.

\subsection{Loss functions}

\textbf{Attention distillation loss. } To distill knowledge based on features for the encoder and depth encoder, this work calculates the attention distillation loss function for these two parts separately. Let $\tilde{\phi}^ C$ denote the $L_2$ channel average pooling function of norm 2, and $\hat{f}^S_n$ represent the student features as the same size as the teacher features after upsampling or downsampling. Let $f^{T,e}_m$ and $f^{S,e}_n$ denote the $m$-th teacher feature and the $n$-th student feature of the encoder, respectively.  ${\alpha}^e_{m,n}$ represents their attention value. Similarly, let $f^{T,d}_m$ and $f^{S,d}_n$ denote the $m$-th teacher feature and the $n$-th student feature of the depth encoder, respectively.  ${\alpha}^d_{m,n}$ represents their attention values. Then, the attention distillation loss function for the entire network can be expressed using Equation (\ref{lad1}), Equation (\ref{lad2}), and Equation (\ref{lada}).
\begin{equation}
    L^{e}_{ad} = \sum^{M}_{m=0}\sum^{N}_{n=0}{\alpha}^e_{m,n}||{\tilde{\phi}}^C(f^{T,e}_m)-{\tilde{\phi}}^C(\hat{f}^{S,e}_n)||_2
    \label{lad1}
\end{equation}
\begin{equation}
    L^{d}_{ad} = \sum^{M}_{m=0}\sum^{N}_{n=0}{\alpha}^d_{m,n}||{\tilde{\phi}}^C(f^{T,d}_m)-{\tilde{\phi}}^C(\hat{f}^{S,d}_n)||_2
    \label{lad2}
\end{equation}
\begin{equation}
    L_{ad} = \frac{1}{2}(L^{e}_{ad}+L^{d}_{ad})
    \label{lada}
\end{equation}

\textbf{Reconstruction loss. } To achieve the image super-resolution task, it is necessary to introduce a reconstruction loss function. This function helps to make the network output closely resemble the ground truth in terms of pixel distance and minimize their disparities. Our network employs the $L_1$ loss function as the reconstruction loss function, which is defined by Equation (\ref{lrec2}).
\begin{equation}
    L_{rec}=||I_{HR}-I_{SR}||_1
    \label{lrec2}
\end{equation}
where $I_{HR}$ and $I_{SR}$ represent the ground truth and the student network output results, respectively.

\textbf{Knowledge distillation loss. }The output distillation loss function is used to encourage the output of the student network to better match teacher network. This paper also uses the $L_1$ loss function for this purpose, which is defined by Equation ( \ref{lrec2}).
\begin{equation}
    L_{kd}=||I^T_{SR}-I_{SR}||_1
    \label{lkd}
\end{equation}
where $I^T_{SR}$ and $I_{SR}$ represent the output results of the teacher network and the student network, respectively.

\section{Experiments}\label{sec5}
\subsection{Training strategies}
All models are implemented and built in Python (version 3.7.9) using PyTorch. The experiments conducts on a Linux system (Ubuntu 16.04), with a GeForce RTX 2080Ti GPU, an Intel Core i7-8700 3.20 GHz CPU, 12GB graphics memory size, and CUDA10.2 version. This study uses the Adam optimizer with an initial learning rate of $lr=2\times10^{-4}$ to train the network. Subsequently, the learning rate is adjusted by a cosine annealing strategy. The input low-resolution image resolution was $112\times80$, the reference image had a resolution of $448\times320$, and the batch size was set to 1. The network is trained on the dataset described in Section 3.5.1, with the loss function parameters $\lambda_{rec}$, $\lambda_{kd}$, and $\lambda_{ad}$ setting to be 1, 0.5, and 0.1, respectively. The teacher model uses pre-trained DSRNet network weights and hasn't undergone any further training. The student model is trained from scratch for a total of 250 epochs, with a training duration of approximately 20 to 25 hours.

This study creates a "Deep Sewer Super-Resolution Dataset," which is produced by using real QV image, including 6000 training images, 6000 reference images, 1050 test images, 1050 test reference images, as well as their corresponding depth map data. This dataset consists of 28200 images in total. All comparison experiments are carried out on this dataset.

\subsection{Compare with state-of-the-art methods}

\begin{figure}[!h]
    \centering
    \subfigure{\includegraphics[width=0.3\linewidth]{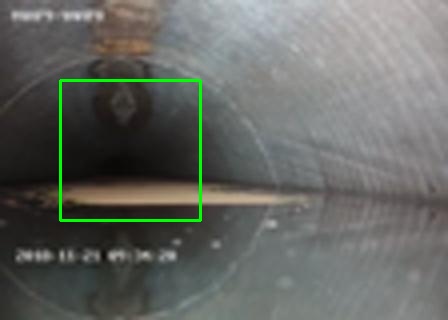}}
    \subfigure{\includegraphics[width=0.3\linewidth]{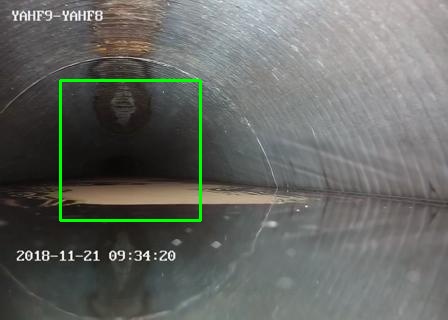}}
    \subfigure{\includegraphics[width=0.3\linewidth]{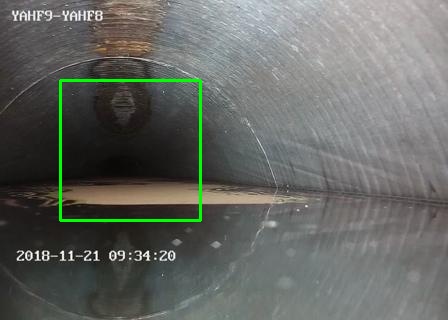}}
    \subfigure{\includegraphics[width=0.3\linewidth]{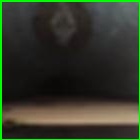}}
    \subfigure{\includegraphics[width=0.3\linewidth]{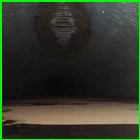}}
    \subfigure{\includegraphics[width=0.3\linewidth]{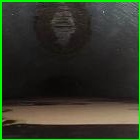}}
    \subfigure{\includegraphics[width=0.3\linewidth]{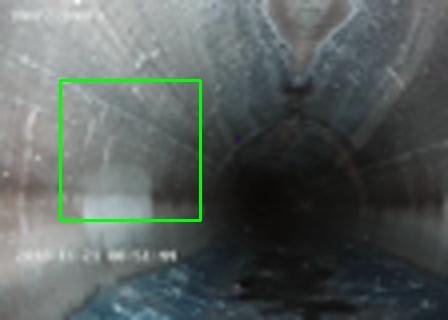}}
    \subfigure{\includegraphics[width=0.3\linewidth]{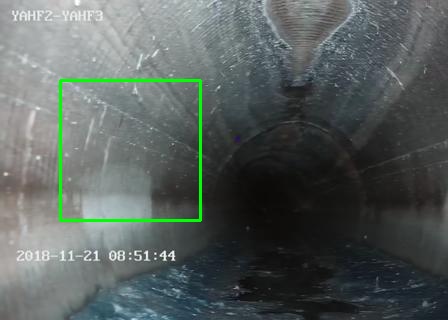}}
    \subfigure{\includegraphics[width=0.3\linewidth]{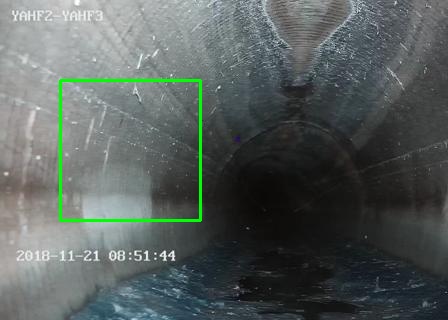}}
    \subfigure{\includegraphics[width=0.3\linewidth]{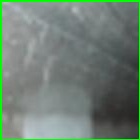}}
    \subfigure{\includegraphics[width=0.3\linewidth]{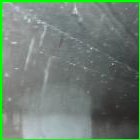}}
    \subfigure{\includegraphics[width=0.3\linewidth]{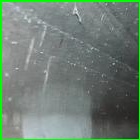}}
    \subfigure{\includegraphics[width=0.3\linewidth]{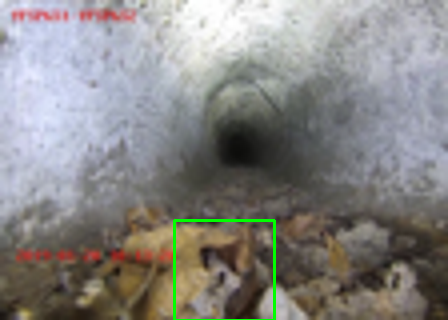}}
    \subfigure{\includegraphics[width=0.3\linewidth]{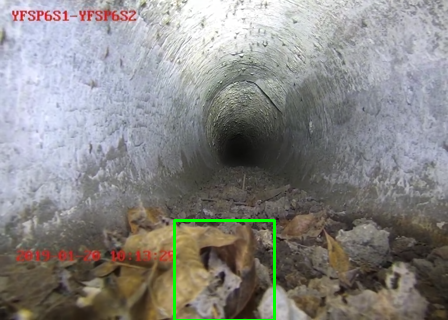}}
    \subfigure{\includegraphics[width=0.3\linewidth]{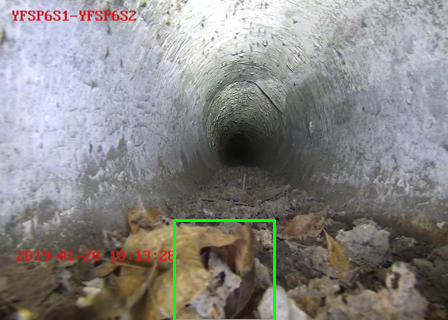}}
    \subfigure{\includegraphics[width=0.3\linewidth]{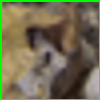}}
    \subfigure{\includegraphics[width=0.3\linewidth]{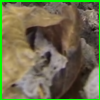}}
    \subfigure{\includegraphics[width=0.3\linewidth]{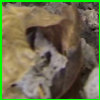}}
    \subfigure{\includegraphics[width=0.3\linewidth]{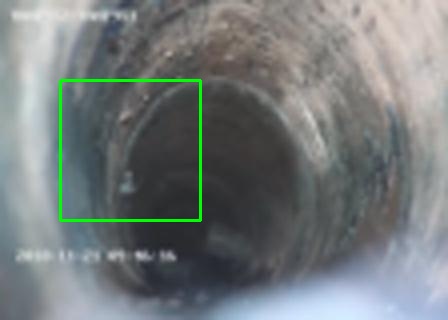}}
    \subfigure{\includegraphics[width=0.3\linewidth]{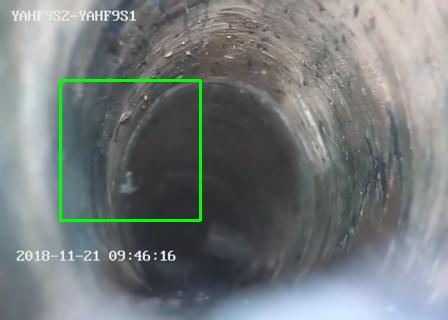}}
    \subfigure{\includegraphics[width=0.3\linewidth]{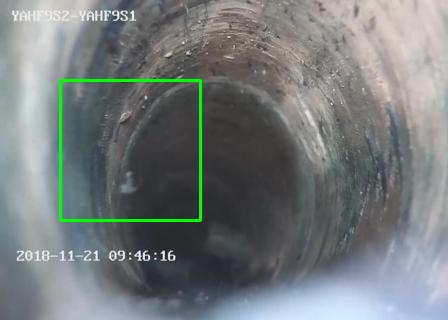}}
    \setcounter{subfigure}{0}
    \subfigure[LR]{\includegraphics[width=0.3\linewidth]{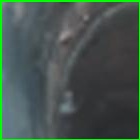}}
    \subfigure[DSRNet]{\includegraphics[width=0.3\linewidth]{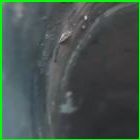}}
    \subfigure[HR]{\includegraphics[width=0.3\linewidth]{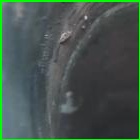}}
    \caption{Experiment results on Deep Sewer Super-Resolution Dataset}
    \label{vis3}
\end{figure}

Figure \ref{vis3} shows the experiment results of image super-resolution. The first column in each figure shows the input low-resolution image, the second column shows the DSRNet output super-resolution image, and the third column shows the high-resolution ground truth. By comparing the results, it can be inferred that the proposed DSRNet effectively rectifies and reconstructs low-resolution images, generating clear texture details and improving overall image clarity.

\paragraph{Evaluation Metrics}
Experiments use Peak Signal-to-Noise Ratio (PSNR) and Structural Similarity index (SSIM) as metrics. Both are widely used metrics in image processing to measure the quality of reconstructed images, specifically in the context of super-resolution tasks. PSNR expressed in decibels (dB) and compare mean-square error of images, while SSIM ranges from -1 to 1, considers changes in structural information, luminance, and contrast.

\paragraph{DSRNet}
To evaluate the effectiveness of the proposed DSRNet, this paper compares it to other existing mainstream image super-resolution methods using the Depth SR dataset. All methods are retrained and tested on this dataset. The methods used for comparison including RCAN~\citep{zhang2018image}, EDSR~\citep{lim2017enhanced}, DBPN~\citep{haris2018deep}, MDSR~\citep{hu2019meta}, ATD~\citep{zhang2024transcending}, RCAN+ with self-supervised strategy enhancement of RCAN, MASA-SR~\citep{lu2021masa}, and MASA-SR trained using only $L_{rec}$, denoted as MASA$-rec$. Among them, ATD is the latest advanced transformer SR method.

\begin{table}[!h]
    \centering
    \caption{Compare with state-of-the-art methods}
    \begin{tabular}{ccccc}
    \toprule[1.5pt]
        method & PSNR & SSIM & FLOPs(G) & Params(M) \\ 
        \midrule[1pt]  
        EDSR & 35.5717 & 0.9153 & 37.39 & 1.81 \\ 
        DBPN & 35.6367 & 0.9163 & 314.81 & 2.21 \\ 
        MDSR & 35.7053 & 0.9158 & 44.11 & 2.22 \\ 
        RCAN  & 35.7180 & 0.9154 & 261.44 & 15.59 \\
        RCAN+ & 36.0771 & 0.9191 & 261.44 & 15.59 \\
        MASA-SR & 36.3563 & 0.9555 & 367.93 & 4.03 \\
        MASA$-rec$ & \underline{36.7557} & \underline{0.9597} & 367.93 & 4.03 \\
        ATD & 35.8824 & 0.9439 & 300.23 & 20.3 \\
        ATD-light & 35.8088 & 0.9425 & 11.41 & 0.769 \\
        DSRNet & 36.2170 & 0.9526 & 379.75 & 8.13 \\
        DSRNet+ & \textbf{36.9621} & \textbf{0.9598} & 379.75 & 8.13 \\
        DSRNet-S & 36.5064 & 0.9554 & 250.81 & 5.03 \\
        DSRNet-D & 36.6797 & 0.9560 & 250.81 & 5.03 \\
        \bottomrule[1.5pt]
    \end{tabular}
    \label{compare1}
\end{table}

\textbf{Quantitative analysis. } Table \ref{compare1} presents the results of the quantitative analysis for the proposed DSRNet and other image super-resolution methods. Yang et al. pointed out that $L_{per}$ and $L_{adv}$ can enhance the visual effect of the image, but they may reduce the values of PSNR and SSIM. Therefore, this work proposes a DSRNet trained only using $L_{rec}$ and $L_{dep}$, which is referred to as DSRNet+. The FLOPs in the table are calculated based on a low-resolution image with a size of $128\times128$. DSRNet+ achieves the best results in both PSNR and SSIM metrics, with values of 36.962126 and 0.959806, respectively. MASA$-rec$ achieves second place in PSNR and SSIM metrics, while other methods perform relatively close. However, overall, their performance was not as good as that of the DSRNet+ method. It should be noted that both MASA-SR and DSRNet in the table are reference-based super-resolution methods, while the other methods are single-image super-resolution methods. Reference-based super-resolution methods have advantages in terms of quantization indicators because they introduce more prior information about reference images compared to single-image super-resolution methods. DSRNet further introduces the prior information from depth maps, which is not available in traditional reference-based super-resolution methods, thus obtaining better results.

Additionally, from the table it can be seen that the student model DSRNet-S has lower PSNR and SSIM compared to the teacher model DSRNet-T, which can be attributed to its lightweight network structure.  However, the DSRNet-D model,  trained using knowledge distillation, improves PSNR and SSIM while maintaining a low model complexity, resulting in good outcomes. Specifically, DSRNet-D improves PSNR by about 0.11 compared to DSRNet-S and SSIM by approximately 0.01, resulting in an improved method performance. Although the performance of the DSRNet-D method is slightly lower than that of the teacher model DSRNet-T, the FLOPs and Params decrease by 128.94 and 3.10, respectively. The speed of the model has been improved, and the size of the model has also been compressed.

\begin{figure}[!h]
    \centering
    \includegraphics[width=0.45\textwidth]{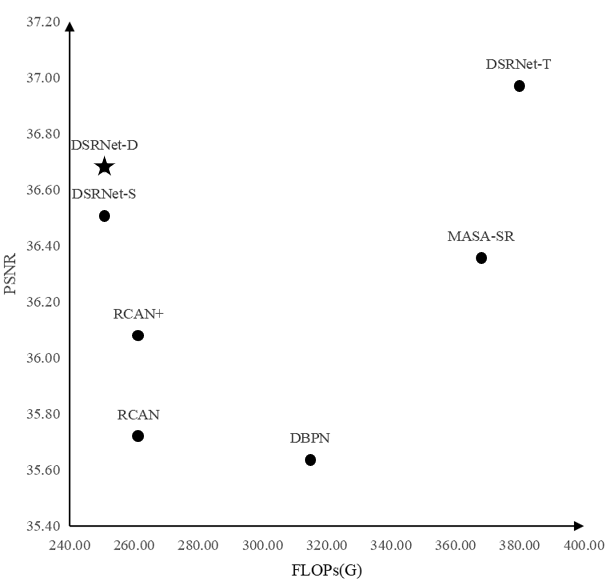}
    \caption{The relation between FLOPs and PSNR}
    \label{flopspsnr}
\end{figure}
Figure \ref{flopspsnr} more intuitively displays the relationship between FLOPs and PSNR of different methods. It can be considered that the method near the upper-left corner of the graph is better in terms of a comprehensive evaluation of speed and effectiveness. It can be seen that DSRNet-D achieves the lowest FLOPs and the second-highest PSNR value in this experiment. This indicates that the knowledge distillation method proposed in this work can effectively accelerate the speed of the network, reduce the number of network parameters, and to some extent compensate for the impact of the lightweight network structure on method performance.

\begin{figure}[!h]
    \centering
    \subfigure[HR 1]{\includegraphics[width=0.4\linewidth]{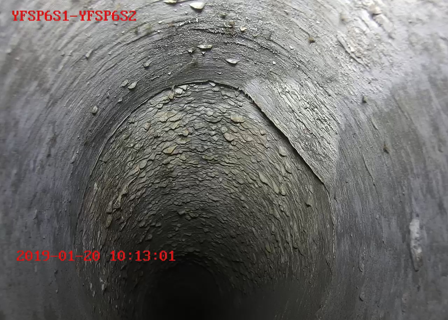}}
    \subfigure[HR 2]{\includegraphics[width=0.4\linewidth]{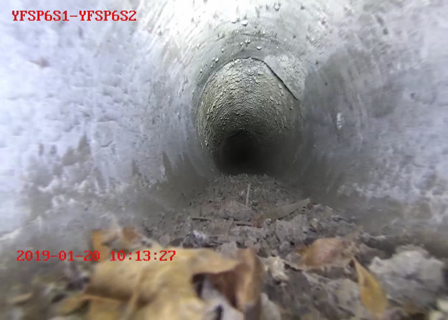}} 
    \subfigure[MASA]{\includegraphics[width=0.3\linewidth]{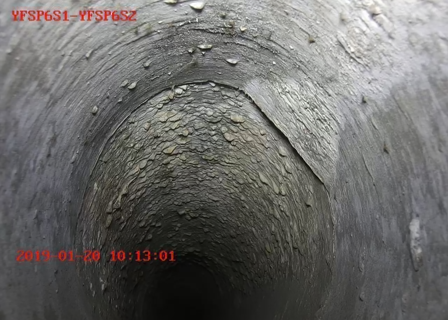}}
    \subfigure[ATD]{\includegraphics[width=0.3\linewidth]{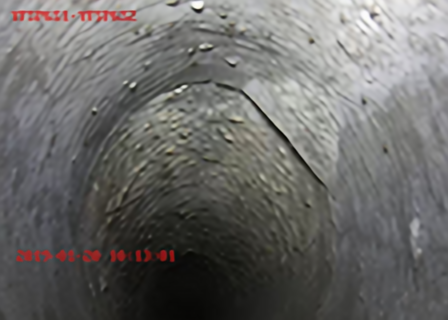}} 
    \subfigure[DSRNet]{\includegraphics[width=0.3\linewidth]{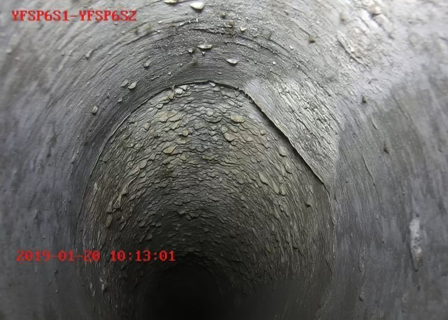}}  
    \subfigure[MASA]{\includegraphics[width=0.3\linewidth]{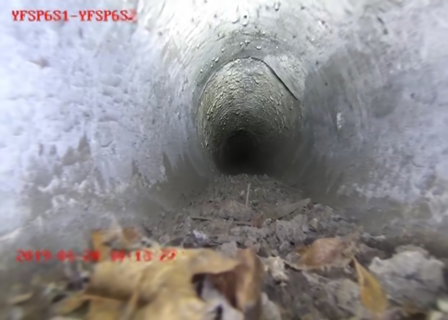}}
    \subfigure[ATD]{\includegraphics[width=0.3\linewidth]{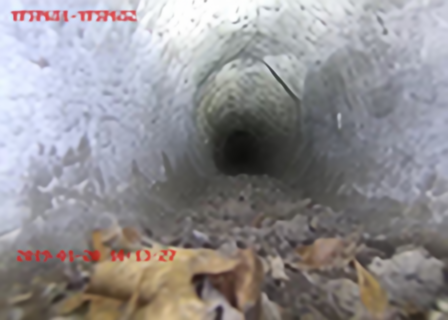}} 
    \subfigure[DSRNet]{\includegraphics[width=0.3\linewidth]{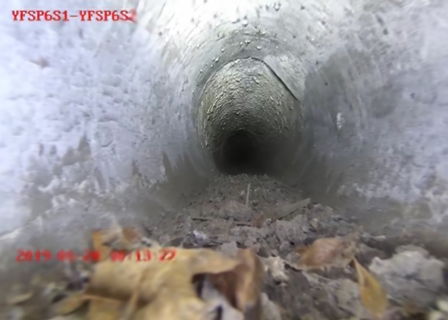}} 
    \caption{Visual results compared with other methods}
    \label{qt1}
\end{figure}

\begin{figure}[!h]
    \centering
    \subfigure[HR]{
    \begin{minipage}[b]{0.6\linewidth}
		\includegraphics[width=\linewidth]{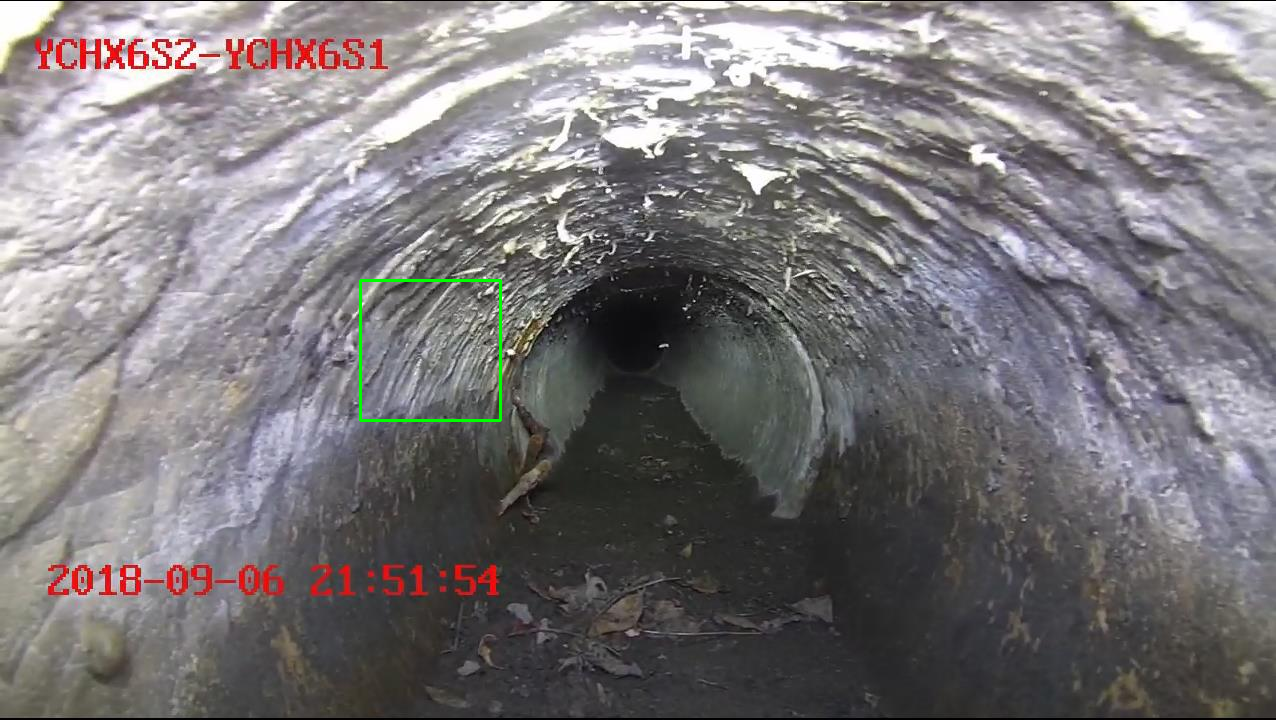}
    \end{minipage}
    }
    
    \subfigure[HR]{\includegraphics[width=0.3\linewidth]{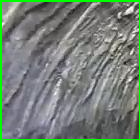}}
    \subfigure[LR]{\includegraphics[width=0.3\linewidth]{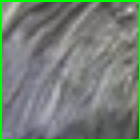}} 
    \subfigure[EDSR]{\includegraphics[width=0.3\linewidth]{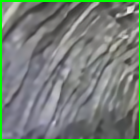}}
    \subfigure[DBPN]{\includegraphics[width=0.3\linewidth]{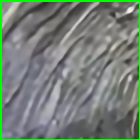}} 
    \subfigure[MDSR]{\includegraphics[width=0.3\linewidth]{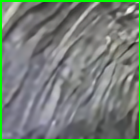}} 
    \subfigure[DSRNet]{\includegraphics[width=0.3\linewidth]{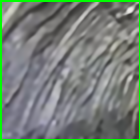}}
    \caption{Visual results compared with other methods}
    \label{qt2}
\end{figure}

\textbf{Qualitative analysis. } Figures \ref{qt1} and \ref{qt2} show a visual comparison of the proposed DSRNet with other prominent methods for single-image super-resolution and reference-based super-resolution. It can be seen that, compared to other super-resolution methods, DSRNet can also achieve better recovery results.

\begin{figure}[!h]
    \centering
    \subfigure[JO LR]{\includegraphics[width=0.3\linewidth]{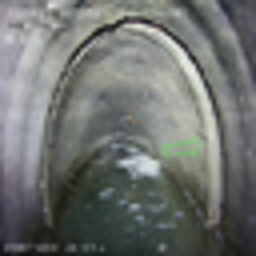}} 
    \subfigure[JO SR]{\includegraphics[width=0.3\linewidth]{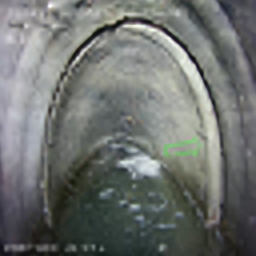}} 
    \subfigure[JO HR]{\includegraphics[width=0.3\linewidth]{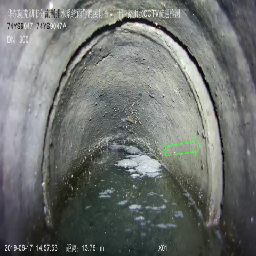}} 
    \subfigure[IL LR]{\includegraphics[width=0.3\linewidth]{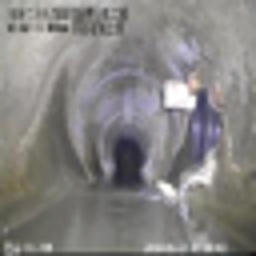}}
    \subfigure[IL SR]{\includegraphics[width=0.3\linewidth]{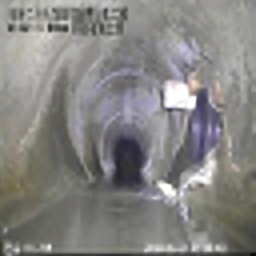}} 
    \subfigure[IL HR]{\includegraphics[width=0.3\linewidth]{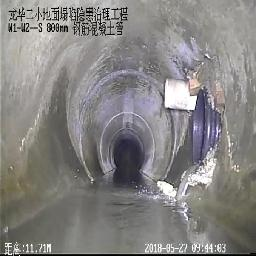}}
    \subfigure[IN LR]{\includegraphics[width=0.3\linewidth]{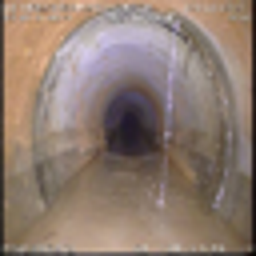}}
    \subfigure[IN SR]{\includegraphics[width=0.3\linewidth]{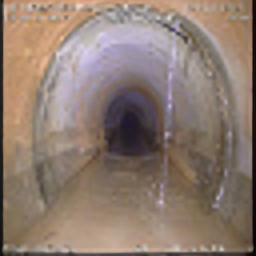}}
    \subfigure[IN HR]{\includegraphics[width=0.3\linewidth]{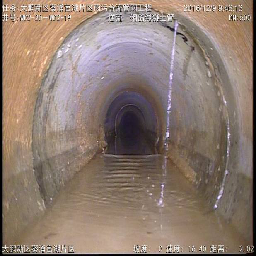}}
    \caption{Defects in Pipe dataset}
    \label{df1}
\end{figure}

\begin{figure}[!h]
    \centering
    \subfigure[IN SR(D)]{\includegraphics[width=0.3\linewidth]{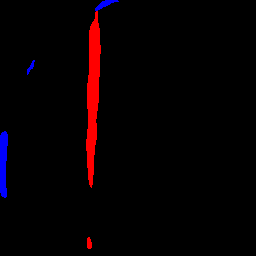}} 
    \subfigure[IN SR(A)]{\includegraphics[width=0.3\linewidth]{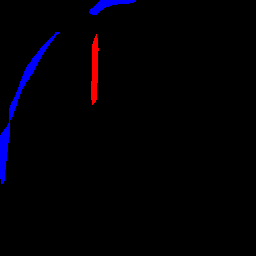}} 
    \subfigure[IN HR]{\includegraphics[width=0.3\linewidth]{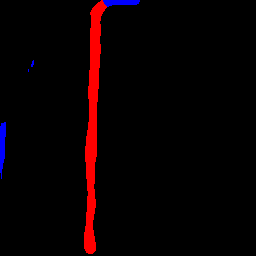}}
    \caption{segmentation in Pipe dataset infiltration class}
    \label{sg1}
\end{figure}

\begin{table*}[h]
    \centering
    \caption{FCN defect semantic segmentation results (mIoU\%)}
    \begin{tabular}{c|cccc|ccccc}
    \toprule[1.5pt]
        Class & LR & SR(D) & SR(A) & HR & ${\Delta}LR$ & ${\Delta}SR$(D) & ${\Delta}SR$(A) & ${\Delta}HR$(D) & ${\Delta}HR$(A) \\ 
        \midrule[1pt]
        JO & 57.90 & 60.49 & 57.74 & 63.36 & -5.47 & 2.59 & -0.16 & 2.87 & 5.62 \\
        IL & 27.02 & 41.42 & 40.08 & 46.01 & -18.99 & 14.40 & 13.06 & 4.60 & 5.93 \\
        IN & 43.69 & 45.69 & 34.71 & 50.94 & -7.24 & 2.00 & -8.98 & 5.24 & 16.23 \\
        Mean & 45.06 & 48.92 & 45.90 & 51.54 & -6.47 & 3.86 & 0.84 & 2.61 & 5.64 \\
        \bottomrule[1.5pt]
    \end{tabular}
    \label{seg1}
\end{table*}

\begin{table*}[h]
    \centering
    \caption{DeepLabv3+ defect semantic segmentation results(mIoU\%)}
    \begin{tabular}{c|cccc|ccccc}
    \toprule[1.5pt]
        Class & LR & SR(D) & SR(A) & HR & ${\Delta}LR$ & ${\Delta}SR$(D) & ${\Delta}SR$(A) & ${\Delta}HR$(D) & ${\Delta}HR$(A) \\ 
        \midrule[1pt]
        JO & 53.39 & 64.53 & 62.46 & 69.96 & -16.57 & 11.14 & 9.07 & 5.43 & 7.50 \\
        IL & 24.57 & 39.71 & 40.41 & 45.30 & -20.73 & 15.14 & 15.84 & 5.59 & 4.89 \\
        IN	& 50.53 & 54.05 & 44.09 & 58.82 & -8.29 & 3.52 & -6.44 & 4.77 & 14.73\\
        Mean & 45.12 & 51.20 & 48.91 & 54.43 & -9.31 & 6.08 & 3.79 & 3.23 & 5.52\\
        \bottomrule[1.5pt]
    \end{tabular}
    \label{seg2}
\end{table*}

\begin{table*}[h]
    \centering
    \caption{YOLOv5s defect localization results (mAP\%)}
    \begin{tabular}{c|cccc|ccccc}
    \toprule[1.5pt]
        Class & LR & SR(D) & SR(A) & HR & ${\Delta}LR$ & ${\Delta}SR$(D) & ${\Delta}SR$(A) & ${\Delta}HR$(D) & ${\Delta}HR$(A) \\ 
        \midrule[1pt]
        JO	& 78.98 & 79.87 & 75.24 & 80.03 & -1.05 & 0.89 & -3.74 & 0.16 & 4.79 \\ 
        IL	& 55.12 & 65.80 & 67.35 & 62.83 & -7.71 & 10.68 & 12.23 & -2.97 & -4.52\\ 
        IN	& 48.25 & 61.38 & 60.31 & 62.95 & -14.70 & 13.13 & 12.06 & 12.96 & 2.64\\ 
        Mean & 60.78 & 69.02 & 67.64 & 68.61 & -8.13 & 8.24 & 6.86 & 3.38 & 0.97\\ 
        \bottomrule[1.5pt]
    \end{tabular}
    \label{ob1}
\end{table*}

\begin{table*}[h]
    \centering
    \caption{SSD defect localization results (mAP\%)}
    \begin{tabular}{c|cccc|ccccc}
    \toprule[1.5pt]
        Class & LR & SR(D) & SR(A) & HR & ${\Delta}LR$ & ${\Delta}SR$(D) & ${\Delta}SR$(A) & ${\Delta}HR$(D) & ${\Delta}HR$(A) \\ 
        \midrule[1pt]
        JO	& 52.59 & 61.08 & 60.90 & 70.10 & -17.51 & 8.49 & 8.31 & 9.02 & 9.20 \\ 
        IL	& 57.73 & 56.91 & 57.82 & 54.23 & 3.50 & -0.82 & 0.09 & -2.68 & -3.59\\ 
        IN	& 28.45 & 43.95 & 42.93 & 51.51 & -23.06 & 15.50 & 14.48 & 7.56 & 8.58\\
        Mean & 46.26 & 53.98 & 53.88 & 58.61 & -12.36 & 7.72 & 7.62 & 4.63 & 4.73\\ 
        \bottomrule[1.5pt]
    \end{tabular}
    \label{ob2}
\end{table*}

\begin{table*}[h]
    \centering
    \caption{GoogleNet InceptionV3 defect classification results (F1\%)}
    \begin{tabular}{c|cccc|ccccc}
    \toprule[1.5pt]
        Class & LR & SR(D) & SR(A) & HR & ${\Delta}LR$ & ${\Delta}SR$(D) & ${\Delta}SR$(A) & ${\Delta}HR$(D) & ${\Delta}HR$(A) \\ 
        \midrule[1pt]
        RB & 30.00 & 27.78 & 27.03 & 30.43 & -0.43 & -2.22 & -2.97 & 2.65 & 3.40 \\ 
        OB & 68.97 & 71.15 & 65.31 & 69.09 & -0.12 & 2.18 & -3.66 & -2.06 & 3.78 \\ 
        PF & 35.29 & 51.43 & 61.11 & 54.55 & -19.26 & 16.14 & 25.82 & 3.12 & -6.56 \\ 
        DE & 54.55 & 61.90 & 62.92 & 66.67 & -12.12 & 7.35 & 8.37 & 4.77 & 3.75 \\ 
        FS & 75.63 & 78.69 & 77.05 & 77.52 & -1.89 & 3.06 & 1.42 & -1.17 & 0.47 \\ 
        IS & 25.00 & 21.43 & 20.00 & 21.05 & 3.95 & -3.57 & -5.00 & -0.38 & 1.05 \\ 
        RO & 33.33 & 38.71 & 40.00 & 35.71 & -2.38 & 5.38 & 6.67 & -3.00 & -4.29 \\ 
        IN & 37.50 & 38.10 & 38.10 & 36.36 & 1.14 & 0.60 & 0.60 & -1.74 & -1.74 \\ 
        AF & 14.81 & 18.18 & 35.90 & 34.78 & -19.97 & 3.37 & 21.09 & 16.60 & -1.12 \\ 
        BE & 45.45 & 46.38 & 45.95 & 50.63 & -5.18 & 0.93 & 0.50 & 4.25 & 4.68 \\ 
        FO & 14.55 & 16.00 & 18.18 & 17.39 & -2.84 & 1.45 & 3.63 & 1.39 & -0.79 \\ 
        GR & 36.84 & 43.75 & 41.18 & 48.28 & -11.44 & 6.91 & 4.34 & 4.53 & 7.10 \\ 
        PH & 66.67 & 57.14 & 50.00 & 44.44 & 22.23 & -9.53 & -16.67 & -12.70 & -5.56 \\ 
        OP & 12.50 & 13.33 & 0.00 & 36.36 & -23.86 & 0.83 & -12.50 & 23.03 & 36.36 \\ 
        OK & 0.00 & 0.00 & 0.00 & 14.29 & -14.29 & 0.00 & 0.00 & 14.29 & 14.29 \\ 
        Mean & 36.74 & 38.93 & 26.67 & 42.50 & -5.76 & 2.19 & -10.07 & 3.57 & 15.83 \\
        \bottomrule[1.5pt]
    \end{tabular}
    \label{cls1}
\end{table*}

\begin{table*}[h]
    \centering
    \caption{ResNet101 defect classification results (F1\%)}
    \begin{tabular}{c|cccc|ccccc}
    \toprule[1.5pt]
        Class & LR & SR(D) & SR(A) & HR & ${\Delta}LR$ & ${\Delta}SR$(D) & ${\Delta}SR$(A) & ${\Delta}HR$(D) & ${\Delta}HR$(A) \\ 
        \midrule[1pt]
        RB & 22.22 & 23.81 & 21.74 & 27.91 & -5.69 & 1.59 & -0.48 & 4.10 & 6.17 \\ 
        OB & 72.90 & 71.15 & 75.93 & 75.23 & -2.33 & -1.75 & 3.03 & 4.08 & -0.70 \\ 
        PF & 53.85 & 60.00 & 54.05 & 55.56 & -1.71 & 6.15 & 0.20 & -4.44 & 1.51 \\ 
        DE & 63.16 & 76.74 & 81.19 & 80.81 & -17.65 & 13.58 & 18.03 & 4.07 & -0.38 \\ 
        FS & 66.67 & 74.58 & 74.60 & 74.24 & -7.57 & 7.91 & 7.93 & -0.34 & -0.36 \\ 
        IS & 8.00 & 16.00 & 20.69 & 22.22 & -14.22 & 8.00 & 12.69 & 6.22 & 1.53 \\ 
        RO & 33.33 & 44.44 & 32.43 & 41.03 & -7.70 & 11.11 & -0.90 & -3.41 & 8.60 \\ 
        IN & 44.44 & 48.48 & 43.24 & 43.90 & 0.54 & 4.04 & -1.20 & -4.58 & 0.66 \\ 
        AF & 10.00 & 20.00 & 34.29 & 33.33 & -23.33 & 10.00 & 24.29 & 13.33 & -0.96 \\ 
        BE & 49.06 & 53.97 & 54.79 & 57.83 & -8.77 & 4.91 & 5.73 & 3.86 & 3.04 \\ 
        FO & 23.53 & 20.00 & 30.00 & 21.05 & 2.48 & -3.53 & 6.47 & 1.05 & -8.95 \\ 
        GR & 51.85 & 57.14 & 50.00 & 51.61 & 0.24 & 5.29 & -1.85 & -5.53 & 1.61 \\ 
        PH & 63.16 & 60.00 & 60.00 & 44.44 & 18.72 & -3.16 & -3.16 & -15.56 & -15.56 \\ 
        OP & 40.00 & 40.00 & 0.00 & 36.36 & 3.64 & 0.00 & -40.00 & -3.64 & 36.36 \\ 
        OK & 46.15 & 54.55 & 0.00 & 54.55 & -8.40 & 8.40 & -46.15 & 0.00 & 54.55 \\ 
        Mean & 43.22 & 48.06 & 30.77 & 48.00 & -4.78 & 4.84 & -12.45 & -0.06 & 17.23 \\
        \bottomrule[1.5pt]
    \end{tabular}
    \label{cls2}
\end{table*}

\begin{table*}[h]
    \centering
    \caption{IDCNN\&NDCNN defect classification results (F1\%)}
    \begin{tabular}{c|cccc|ccccc}
    \toprule[1.5pt]
        Class & LR & SR(D) & SR(A) & HR & ${\Delta}LR$ & ${\Delta}SR$(D) & ${\Delta}SR$(A) & ${\Delta}HR$(D) & ${\Delta}HR$(A) \\ 
        \midrule[1pt]
        RB & 12.77 & 20.25 & 24.00 & 22.86 & -10.09 & 7.48 & 11.23 & 2.61 & -1.14 \\ 
        OB & 38.24 & 65.38 & 63.49 & 61.07 & -22.83 & 27.14 & 25.25 & -4.31 & -2.42 \\ 
        PF & 21.57 & 30.77 & 31.37 & 40.74 & -19.17 & 9.20 & 9.80 & 9.97 & 9.37 \\ 
        DE & 27.91 & 34.62 & 31.71 & 36.59 & -8.68 & 6.71 & 3.80 & 1.97 & 4.88 \\ 
        FS & 40.43 & 64.52 & 69.39 & 67.55 & -27.12 & 24.09 & 28.96 & 3.03 & -1.84 \\ 
        IS & 6.90 & 12.31 & 9.64 & 11.76 & -4.86 & 5.41 & 2.74 & -0.55 & 2.12 \\ 
        RO & 7.14 & 22.22 & 16.28 & 15.22 & -8.08 & 15.08 & 9.14 & -7.00 & -1.06 \\ 
        IN & 28.57 & 19.05 & 19.61 & 16.67 & 11.90 & -9.52 & -8.96 & -2.38 & -2.94 \\ 
        AF & 11.43 & 11.43 & 8.51 & 8.33 & 3.10 & 0.00 & -2.92 & -3.10 & -0.18 \\ 
        BE & 40.68 & 38.71 & 37.50 & 33.04 & 7.64 & -1.97 & -3.18 & -5.67 & -4.46 \\ 
        FO & 8.16 & 6.84 & 6.72 & 6.45 & 1.71 & -1.32 & -1.44 & -0.39 & -0.27 \\ 
        GR & 26.67 & 16.00 & 16.84 & 16.00 & 10.67 & -10.67 & -9.83 & 0.00 & -0.84 \\ 
        PH & 0.00 & 11.11 & 15.19 & 15.00 & -15.00 & 11.11 & 15.19 & 3.89 & -0.19 \\ 
        OP & 2.63 & 3.45 & 0.00 & 4.88 & -2.25 & 0.82 & -2.63 & 1.43 & 4.88 \\ 
        OK & 8.82 & 6.74 & 0.00 & 7.69 & 1.13 & -2.08 & -8.82 & 0.95 & 7.69 \\ 
        Mean & 18.79 & 24.23 & 5.71 & 24.26 & -5.47 & 5.44 & -13.08 & 0.03 & 18.55 \\ 
        \bottomrule[1.5pt]
    \end{tabular}
    \label{cls3}
\end{table*}

\subsection{Application in the Pipe and Sewer-ML dataset}
To verify the effectiveness of the proposed DSRNet in enhancing the performance of subsequent computer vision tasks, three different sewer defect detection experiments were conducted. These experiments included semantic segmentation, object detection, and image classification. This paper uses raw high-resolution images (HR), downsampled low-resolution images (LR), and DSRNet to process super-resolution images (SR) for these experiments. Due to the lack of reference images in the dataset used in this part of the experiment, our work refers to method by \cite{yang2020learning} and uses downsampled low-resolution images as reference images for image super-resolution processing. Since no additional information was obtained from the reference images, DSRNet is essentially a single-image super-resolution method in this case. We still compare it with the state-of-the-art method, ATD.

To demonstrate the improvement of super-resolution networks on advanced visual tasks, in addition to commonly using evaluation indicators for specific tasks, three evaluation indicators: ${\Delta}LR$, ${\Delta}SR$, and ${\Delta}HR$ are proposed in this work. Their calculation formulas are shown in Equation (\ref{dlr}), Equation (\ref{dsr}), and Equation (\ref{dhr}). 

${\Delta}LR$ represents the difference between the downsampled low-resolution images and the high-resolution images. ${\Delta}SR$ represents the difference between the super-resolution images processed by DSRNet and the high-resolution images. ${\Delta}HR$ represents the difference between the high-resolution images and the super-resolution images. These indicators can reflect the performance improvement of DSRNet in terms of image quality and restoration accuracy. 

\begin{equation}
    {\Delta}LR = LR - HR
    \label{dlr}
\end{equation}
\begin{equation}
    {\Delta}SR = SR - LR
    \label{dsr}
\end{equation}
\begin{equation}
    {\Delta}HR = HR - SR
    \label{dhr}
\end{equation}

Here, $LR$, $SR$, and $HR$ represent the evaluation index values obtained on specific tasks for low-resolution images, DSRNet output super-resolution images, and original high-resolution images, respectively. ${\Delta}LR$ is usually a negative number, and the higher the absolute value of this value, the greater the degree of decrease caused by low resolution on a specific indicator. ${\Delta}SR$ reflects the degree of improvement of DSRNet relative to low-resolution images on specific indicators, usually a positive number. A larger value indicates better performance of DSRNet for specific tasks. ${\Delta}HR$ reflects the difference between DSRNet and high-resolution images in specific indicators, usually a positive number. The smaller the value, the smaller the difference between DSRNet's results on a specific task and the ideal situation. If this value is negative, it indicates that the output super-resolution image of DSRNet performs better on a specific task than the original high-resolution image.

\paragraph{Sewer defect semantic segmentation}
The data for the semantic segmentation experiment is from our previous work ~\citep{pan2020Automatic}, which consists of images with sizes of $256\times256$. The dataset contains three kinds of defects, as shown in figure \ref{df1}, where JO stands for Joint offset, IL for Intruding lateral and IN for Infiltration. Two mainstream semantic segmentation methods are used for defect semantic segmentation experiments: FCN and DeepLabv3+. The commonly used evaluation indicator for semantic segmentation tasks is Mean Intersection over Union (mIoU). Tables \ref{seg1} and \ref{seg2} show the experiment results of semantic segmentation tasks using FCN and DeepLabv3+, respectively. The DSRNet and ATD~\citep{zhang2024transcending} were tested simultaneously, with the results obtained by DSRNet labeled (D) and those obtained by ATD labeled (A), the meanings of suffixes appearing in subsequent experiments remains the same. From the two tables, it can be seen that the ${\Delta}SR$ of FCN is 0.0386, and the ${\Delta}SR$ of DeepLabv3+ is 0.0608, indicating that the DSRNet proposed in this paper has a significant improvement in sewer defect semantic segmentation tasks. It can be seen that ATD also has an effect on low-resolution images, but in specific tasks, it performs similarly to or worse than DSRNet, especially in the infiltration classification where it even has a negative impact, figure \ref{sg1} shows one example. It is worth noting that the improvement of resolution is significant for segmentation of class IL. Compared with the other classes, class IL has the smallest ${\Delta}LR$ and the largest ${\Delta}SR$. 

\paragraph{Sewer defect localization}
The object detection experiment also used the Pipe dataset of \cite{pan2020Automatic}. Two object detection methods are used for defect object detection experiments in sewer scenarios: YOLOv5s and SSD. The commonly used evaluation indicator for object detection tasks is mean Average Precision (mAP). Tables \ref{ob1} and \ref{ob2} show the experiment results of using YOLOv5s and SSD for object detection tasks, respectively. It can be seen that the IN class is most sensitive to resolution in localization tasks, and the ${\Delta}SR$ of YOLOv5s is 0.0824, and the ${\Delta}SR$ of SSD is 0.0772, both 
are positive numbers. Therefore, it can be concluded that the DSRNet proposed in this work can effectively improve the efficiency of sewer defect target detection tasks. However, DSRNet also has limitations, as the improvement it brings to the IL class is slightly less compared to the other two classes.

\paragraph{Sewer defect classification}
The image classification experiment uses the Sewer-ML dataset ~\citep{haurum2021sewer}, which is a multi-classification dataset for sewer defects. Three different image classification methods are used for defect classification experiments in sewer scenarios: Google Net Inception V3, ResNet-101, and IDCNN\&NDCNN. Among them, IDCNN\&NDCNN is a network used for sewer defect classification, while the other two networks are image classification networks in the general field. This study uses F1 score to evaluate the performance of image classification networks. Tables \ref{cls1}, \ref{cls2}, and \ref{cls3} show the results of the sewer defect classification experiment. The table reveals that the $\Delta{SR}$ values for the three models are 0.0219, 0.0484, and 0.0544, respectively. This indicates that DSRNet also has a significant improvement in sewer defect classification tasks. It is worth noting that the ${\Delta}HR$ of the defect classification task using ResNet101 is negative, indicating that in this experiment, the images processed by DSRNet were better than those directly using high-resolution images. In the classification task, the limitation of DSRNet is the significantly lower recognition accuracy for the AF (Settled deposits) class after super-resolution compared to ATD. Additionally, the PH (Chiseled connection) class shows an abnormal trend, with accuracy notably increasing as the resolution decreases. The recognition accuracy of OP(Connection with transition profile) and OK(Connection with construction changes) class is seriously affected by the resolution, lead to unrecognized cases in LR and SR.

\subsection{Ablation studies}

\begin{table}[h]
    \centering
    \caption{Ablation study on DSRNet}
    \begin{tabular}{cccccc}
    \toprule[1.5pt]
        $L_{rec}$ & Depth gt & $L_{dep}$ & PSNR & SSIM \\ 
        \midrule[1pt]
        $\checkmark$ &   &   & 36.755700 & 0.959752  \\ 
        $\checkmark$ & $\checkmark$ &   & 36.969101 & 0.959903 \\ 
        $\checkmark$ &  & $\checkmark$ & 36.962126 & 0.959806 \\ 
        \bottomrule[1.5pt]
    \end{tabular}
    \label{ab1}
\end{table}

\begin{table}[h]
    \centering
    \caption{Ablation study on KD and AD Loss}
    \begin{tabular}{ccccc}
    \toprule[1.5pt]
        $L_{rec}$ & $L_{kd}$ & $L_{ad}$ & PSNR & SSIM \\ 
        \midrule[1pt]
        $\checkmark$ &   &   & 36.506430 & 0.955479  \\  
        $\checkmark$ & $\checkmark$ &   & 36.663566 & 0.955925 \\ 
        $\checkmark$ &   & $\checkmark$ & 36.541040 & 0.955604 \\ 
        $\checkmark$ & $\checkmark$ & $\checkmark$ & \textbf{36.679796} & \textbf{0.956011} \\
        \bottomrule[1.5pt]
    \end{tabular}
    \label{ab2}
\end{table}

The key to this work is to demonstrate that the introduction of depth information can contribute to the restoration effect of image super-resolution networks. The previous chapter explains through theoretical analysis how depth extraction networks and DMM affect the reconstruction process of super-resolution methods. In order to show the impact of the depth map and depth reconstruction loss function on the image super-resolution results, ablation experiments are carried out and the results are quantitatively evaluated.

Table \ref{ab1} shows the impact of $L_{dep}$ on the experiment results. The first row indicates using only $L_{rec}$ for the experiment, the second row indicates using the depth ground truth for the experiment, and the third row indicates using the depth extraction network and $L_{dep}$ for the experiment. It can be seen from the table that using only $L_{rec}$ can ensure the image restoration effect and achieve a level of parity with other methods in terms of indicators. The second experiment shows that adding depth ground truths to the image super-resolution network can effectively improve performance, with improvements of 0.2134 and 0.0002 on PSNR and SSIM, respectively, which verifies the effectiveness of the depth map and DMM modules. The third experiment adds the depth extraction network and the corresponding loss function $L_{dep}$. Compared to using depth ground truths, our experiment only differs by 0.0070 and 0.0001 on PSNR and SSIM, indicating that the depth extraction network and $L_{dep}$ can effectively extract depth information. Under the current accuracy, PSNR and SSIM float below the second row, while much higher indicators than that using only $L_{rec}$. The depth extraction network and $L_{dep}$ can also improve the performance of image super-resolution networks.

In order to verify the effect of each loss function, different loss functions are used to train the distillation model. Table \ref{ab2} shows the influence of different loss functions on distillation effect. The first row is the training result of using high-resolution images instead of distillation strategy training, the second row is the result of using high-resolution images and teachers' output monitoring, the third row is the result of using high-resolution images and teachers' features monitoring, and the fourth row is the result of using all training strategies. The table illustrates that upon adding $L_{ad}$ and $L_{kd}$, both PSNR and SSIM of the student model show improvement. 

\section{Conclusion}\label{sec6}
This work introduces super-resolution into sewer defects detection for the first time by proposing a super-resolution network DSRNet based on depth prior and reference images. The joint training of depth estimation network and image super-resolution network is conducted. Besides, the proposed DMM extracts depth features through the depth encoder, and then fuses the depth map features with the features of low resolution images and reference images to achieve a super-resolution method with depth priors. Experiments have proved the effectiveness of DSRNet and the significant improvements that DSRNet brings to downstream pipeline defect localization, segmentation and classification tasks.

A knowledge distillation model on the DSRNet is also conducted by this work. It is based on feature matching, introducing attention mechanism into the feature selection process of knowledge distillation. Experiments show that the distillation model can effectively reduce the number of network parameters and FLOPs while ensuring the accuracy, achieving the balance between performance and speed. The proposed method not only improves the visual effect of low-resolution images but also enhances the performance of subsequent visual tasks, demonstrating its significant application value.

Although super-resolution has significantly improved overall performance in downstream tasks, experiments have shown that certain types of defects are more sensitive to resolution, and for some defects, recognition accuracy increases as resolution decreases. The specific reasons for this and possible solutions require further investigation. Lightweight methods need further exploration to determine whether there are more effective approaches to maximize the reduction of FLOPs and Params while maintaining accuracy.

\bibliography{reference}%

\end{document}